\documentclass[conference]{IEEEtran}
\usepackage{graphicx}
\usepackage{float}
\usepackage{times}
\usepackage{dsfont}
\usepackage{amsmath,amssymb}
\usepackage[numbers]{natbib}
\usepackage{multicol}
\usepackage{siunitx}
\usepackage[bookmarks=true]{hyperref}
\usepackage{cleveref}
\usepackage{xspace}
\usepackage{xurl}
\usepackage{capt-of}
\usepackage[usenames,dvipsnames]{xcolor}
\usepackage{booktabs}  
\usepackage{threeparttable}
\usepackage{multirow}
\usepackage{ulem}

\usepackage{colortbl}
\usepackage{xcolor}
\usepackage{svg}

\definecolor{verylightgray}{RGB}{240, 240, 240}
\definecolor{llnvgreen}{RGB}{220, 237, 191}

\definecolor{rqblue}{RGB}{246, 208, 208}

\definecolor{gsred}{RGB}{201, 201, 247}

\definecolor{mydarkblue}{rgb}{0,0.08,0.45}
\definecolor{mydarkgreen}{RGB}{0, 139, 69}
\definecolor{mygreen2}{RGB}{0 205 0}
\definecolor{mybrown}{RGB}{139 69 19}
\definecolor{Methodred}{RGB}{191, 3, 3} 
\hypersetup{
    colorlinks=true,
    linkcolor=magenta,
    urlcolor=magenta,
    citecolor=mygreen2,
}

\newcommand{\goodnumber}[1]{{\color{Methodred}\textbf{#1}}}

\newcommand{\method}{{\color{Methodred}ASAP}\xspace}

\newcommand{\simtoreal}{{sim-to-real}\xspace}
\newcommand{\bs}[1]{\boldsymbol{#1}}

\newcommand{\dofpos}{{\bs{{q}}_{t}}}

\newcommand{\rootangvel}{{\bs{\omega}^{root}_{t}}}
\newcommand{\gravity}{{\bs{g}_{t}}}

\newcommand{\dofvel}{{\bs{\dot{q}}_{t}}}

\newcommand{\actionprev}{{\bs{a}_{t-1}}}

\newcommand{\dofposhist}{{\bs{{q}}_{t-4:t}}}
\newcommand{\rootangvelhist}{{\bs{\omega}^{root}_{t-4:t}}}
\newcommand{\gravityhist}{{\bs{g}_{t-4:t}}}

\newcommand{\dofvelhist}{{\bs{\dot{q}}_{t-4:t}}}

\newcommand{\actionhist}{{\bs{a}_{t-5:t-1}}}

\begin{document}

\title{ASAP: Aligning Simulation and Real-World Physics \\ for Learning Agile Humanoid Whole-Body Skills}



\author{\authorblockN{Tairan He\textsuperscript{\dag1,2}
\quad Jiawei Gao\textsuperscript{\dag1}
\quad Wenli Xiao\textsuperscript{\dag1,2}
\quad Yuanhang Zhang\textsuperscript{\dag1} \quad Zi Wang\textsuperscript{1} \quad Jiashun Wang\textsuperscript{1} \\ \quad Zhengyi Luo\textsuperscript{1,2} \quad Guanqi He\textsuperscript{1} \quad Nikhil Sobanbabu\textsuperscript{1} \quad Chaoyi Pan\textsuperscript{1} \quad Zeji Yi\textsuperscript{1} \quad Guannan Qu\textsuperscript{1} \quad \\  Kris Kitani\textsuperscript{1} \quad   Jessica Hodgins\textsuperscript{1} \quad   Linxi ``Jim" Fan\textsuperscript{2} \quad Yuke Zhu\textsuperscript{2} \quad Changliu Liu\textsuperscript{1} \quad Guanya Shi\textsuperscript{1}}
\authorblockA{
\textsuperscript{1}Carnegie Mellon University \quad \textsuperscript{2}NVIDIA \quad \textsuperscript{\dag}Equal Contributions \\
Page: \href{https://agile.human2humanoid.com}{\texttt{https://agile.human2humanoid.com}} \quad Code: \href{https://github.com/LeCAR-Lab/ASAP}{\texttt{https://github.com/LeCAR-Lab/ASAP}}
}
}


%

\makeatletter
\let\@oldmaketitle\@maketitle
    \renewcommand{\@maketitle}{\@oldmaketitle
    \centering
    \includegraphics[width=1.0\textwidth]{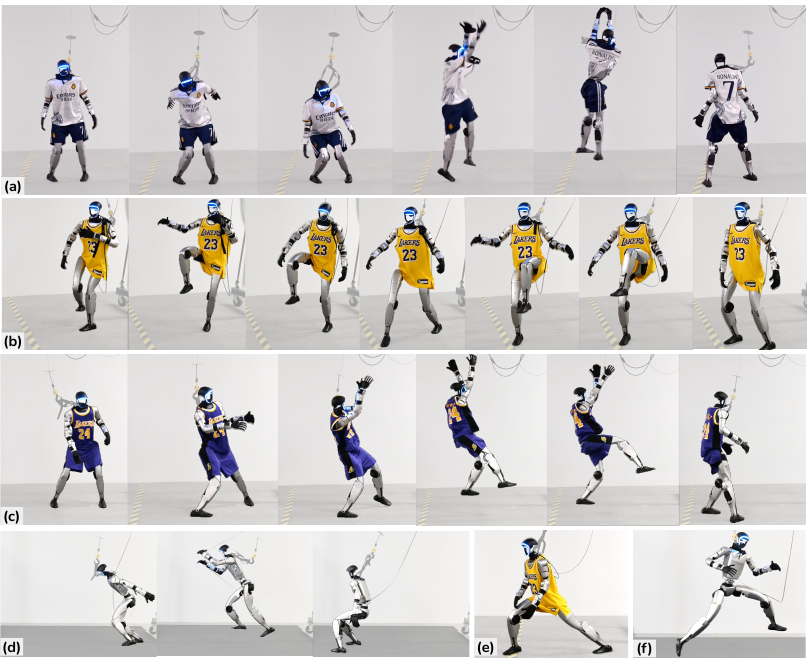}
    
    \vspace{-0.2cm}
    \captionof{figure}{The humanoid robot (Unitree G1) demonstrates diverse agile whole-body skills, showcasing the control policies' agility: (a) Cristiano Ronaldo’s signature celebration involving a jump with a 180-degree mid-air rotation; (b) LeBron James’s ``Silencer'' celebration involving single-leg balancing; and (c) Kobe Bryant’s famous fadeaway jump shot involving single-leg jumping and landing; (d) 1.5m-forward jumping; (e) Leg stretching; (f) 1.3m-side jumping.
    }
    \vspace{-0.3cm} 
    \label{fig:firstpage}
    \setcounter{figure}{1}
  }
\makeatother

\maketitle

\begin{abstract}
\begin{figure*}[t]
    \centering
    \includegraphics[width=\textwidth]{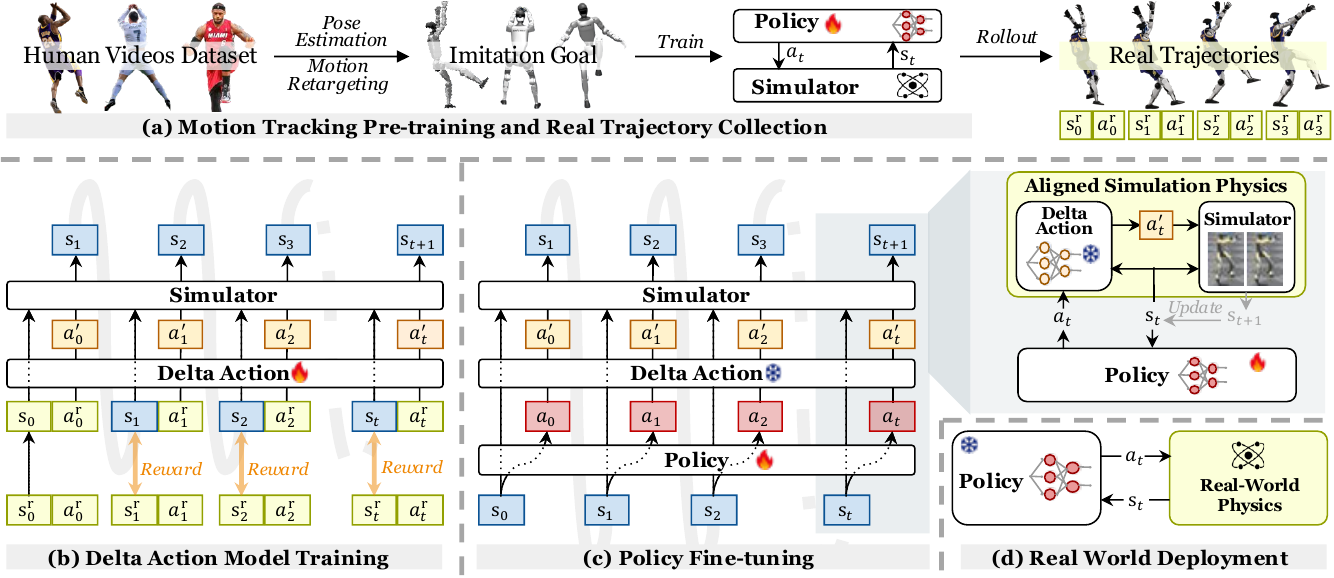}
    \vspace{-2mm}
    \caption{Overview of \method. (a) \textbf{Motion Tracking Pre-training and Real Trajectory Collection}: With the humanoid motions retargeted from human videos, we pre-train multiple motion tracking policies to roll out real-world trajectories. (b) \textbf{Delta Action Model Training}: Based on the real-world rollout data, we train the delta action model by minimizing the discrepancy between simulation state $s_t$ and real-world state $s^r_t$. (c) \textbf{Policy Fine-tuning}: We freeze the delta action model, incorporate it into the simulator to align the real-world physics and then fine-tune the pre-trained motion tracking policy. (d) \textbf{Real-World Deployment}: Finally, we deploy the fine-tuned policy directly in the real world without delta action model.
    }
    \label{fig:ASAP}
    \vspace{-4mm}
\end{figure*}
Humanoid robots hold the potential for unparalleled versatility for performing human-like, whole-body skills. However, achieving agile and coordinated whole-body motions remains a significant challenge due to the dynamics mismatch between simulation and the real world. Existing approaches, such as system identification (SysID) and domain randomization (DR) methods, often rely on labor-intensive parameter tuning or result in overly conservative policies that sacrifice agility. In this paper, we present \method ({\color{Methodred}A}ligning {\color{Methodred}S}imulation and Re{\color{Methodred}a}l {\color{Methodred}P}hysics), a two-stage framework designed to tackle the dynamics mismatch and enable agile humanoid whole-body skills. 
In the first stage, we pre-train motion tracking policies in simulation using retargeted human motion data. In the second stage, we deploy the policies in the real world and collect real-world data to train a delta (residual) action model that compensates for the dynamics mismatch. Then \method fine-tunes pre-trained policies with the delta action model integrated into the simulator to align effectively with real-world dynamics. We evaluate \method across three transfer scenarios—IsaacGym to IsaacSim, IsaacGym to Genesis, and IsaacGym to the real-world Unitree G1 humanoid robot. Our approach significantly improves agility and whole-body coordination across various dynamic motions, reducing tracking error compared to SysID, DR, and delta dynamics learning baselines. 
\method enables highly agile motions that were previously difficult to achieve, demonstrating the potential of delta action learning in bridging simulation and real-world dynamics. These results suggest a promising sim-to-real direction for developing more expressive and agile humanoids.
\end{abstract}

\IEEEpeerreviewmaketitle


\section{Introduction}
\label{sec:introduction}

\begin{figure*}[t]
    \centering
    \includegraphics[width=0.80\textwidth]{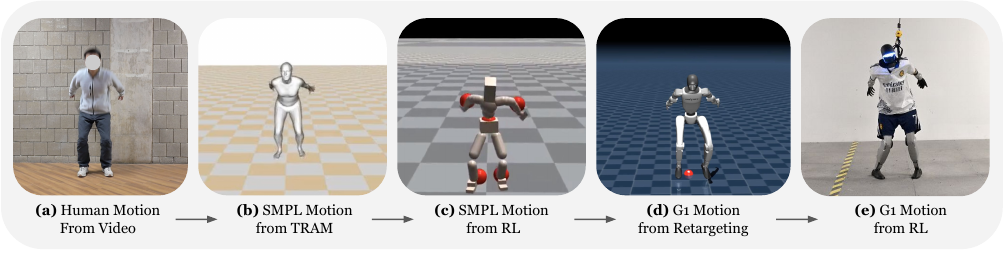}
    \vspace{-3mm}
    \caption{Retargeting Human Video Motions to Robot Motions: (a) Human motions are captured from video. (b) Using TRAM~\cite{wang2025tram}, 3D human motion is reconstructed in the SMPL parameter format. (c) A reinforcement learning (RL) policy is trained in simulation to track the SMPL motion. (d) The learned SMPL motion is retargeted to the Unitree G1 humanoid robot in simulation. (e) The trained RL policy is deployed on the real robot, executing the final motion in the physical world. This pipeline ensures the retargeted motions remain physically feasible and suitable for real-world deployment.}
    \vspace{-3mm}
    \label{fig:data_processing}
\end{figure*}

For decades, we have envisioned humanoid robots achieving or even surpassing human-level agility. However, most prior work~\cite{li2023robust,radosavovic2024real,li2024reinforcement,radosavovic2402humanoid,zhuang2024humanoid,gu2024advancing,Wolf2025AIHumanoid,long2024learning} has primarily focused on locomotion, treating the legs as a means of mobility. Recent studies~\cite{cheng2024expressive,he2024learning,he2024omnih2o,he2024hover,ji2024exbody2} have introduced whole-body expressiveness in humanoid robots, but these efforts have primarily focused on upper-body motions and have yet to achieve the agility seen in human movement. Achieving agile, whole-body skills in humanoid robots remains a fundamental challenge due to not only hardware limits but also the mismatch between simulated dynamics and real-world physics.

Three main approaches have emerged to bridge the dynamics mismatch: System Identification (SysID) methods, domain randomization (DR), and learned dynamics methods. SysID methods directly estimate critical physical parameters, such as motor response characteristics, the mass of each robot link, and terrain properties~\cite{yu2019sim, gu2024advancing}. However, these methods require a pre-defined parameter space~\cite{ljung1998system}, which may not fully capture the sim-to-real gap, especially when real-world dynamics fall outside the modeled distribution. SysID also often relies on ground truth torque measurements~\cite{hwangbo2019learning}, which are unavailable on many widely used hardware platforms, limiting its practical applicability.
DR methods, in contrast, first train control policies in simulation before deploying them on real-world hardwares~\cite{tan2018sim, rudin2022learning, margolis2022rapid}. To mitigate the dynamics mismatch between simulation and real-world physics, DR methods rely on randomizing simulation parameters~\cite{tobin2017domain,peng2018sim}; but this can lead to overly conservative policies~\cite{he2024learning}, ultimately hindering the development of highly agile skills. Another approach to bridge dynamics mismatch is learning a dynamics model of real-world physics using real-world data. While this approach has demonstrated success in low-dimensional systems such as drones~\cite{shi2019neural} and ground vehicles~\cite{xiao2024anycar}, its effectiveness for humanoid robots remains unexplored.

To this end, we propose \method, a two-stage framework that aligns the dynamics mismatch between simulation and real-world physics, enabling agile humanoid whole-body skills. \method involves a pre-training stage where we train base policies in simulation and a post-training stage that finetunes the policy by aligning simulation and real-world dynamics. 
In the \textbf{pre-training} stage, we train a motion tracking policy in simulation using human motion videos as data sources. These motions are first retargeted to humanoid robots~\cite{he2024learning}, and a phase-conditioned motion tracking policy~\cite{peng2018deepmimic} is trained to follow the retargeted movements. However, directly deploying this policy on real hardware results in degraded performance due to the dynamics mismatch.
To address this, the \textbf{post-training} stage collects real-world rollout data, including proprioceptive states and positions recorded by the motion capture system.
The collected data are then replayed in simulation, where the dynamics mismatch manifests as tracking errors. We then train a delta action model that learns to compensate for these discrepancies by minimizing the difference between real-world and simulated states. This model effectively serves as a residual correction term for the dynamics gap. Finally, we fine-tune the pre-trained policy using the delta action model, allowing it to adapt effectively to real-world physics.

We validate \method on diverse agile motions and successfully achieve whole-body agility on the Unitree G1 humanoid robot~\cite{Unitree2024G1}. Our approach significantly reduces motion tracking error compared to prior SysID, DR, and delta dynamics learning baselines in both sim-to-sim (IsaacGym to IsaacSim, IsaacGym to Genesis) and \simtoreal (IsaacGym to Real) transfer scenarios.
Our contributions are summarized below.
\begin{enumerate}
    \item We introduce \method, a framework that bridges the \simtoreal gap by leveraging a delta action model trained via reinforcement learning (RL) with real-world data.
    \item We successfully deploy RL-based whole-body control policies in the real world, achieving previously difficult-to-achieve humanoid motions.
    \item Extensive experiments in both simulation and real-world settings demonstrate that \method effectively reduces dynamics mismatch, enabling highly agile motions on robots and significantly reducing motion tracking errors.
    \item To facilitate smooth transfer between simulators, we develop and open-source a multi-simulator training and evaluation codebase for help accelerate further research.
\end{enumerate}

\section{Pre-training: Learning Agile Humanoid Skills}
\label{sec:deepmimic}

\subsection{Data Generation: Retargeting Human Video Data}
To track expressive and agile motions, we collect a video dataset of human movements and retarget it to robot motions, creating imitation goals for motion-tracking policies, as shown in \Cref{fig:data_processing} and \Cref{fig:ASAP} (a).

\paragraph{Transforming Human Video to SMPL Motions}

We begin by recording videos (see \Cref{fig:data_processing} (a) and \Cref{fig:action_noise}) of humans performing expressive and agile motions. Using TRAM~\cite{wang2025tram}, we reconstruct 3D motions from videos. TRAM estimates the global trajectory of the human motions in SMPL parameter format~\cite{loper2023smpl}, which includes global root translation, orientation, body poses, and shape parameters, as shown in \Cref{fig:data_processing} (b). The resulting motions are denoted as ${\mathcal{D}}_{\text{SMPL}}$.

\paragraph{Simulation-based Data Cleaning}

Since the reconstruction process can introduce noise and errors~\cite{he2024learning}, some estimated motions may not be physically feasible, making them unsuitable for motion tracking in the real world. To address this, we employ a ``sim-to-data'' cleaning procedure. Specifically, we use MaskedMimic~\cite{tessler2024maskedmimic}, a physics-based motion tracker, to imitate the SMPL motions from TRAM in IsaacGym simulator~\cite{makoviychuk2021isaac}. The motions (\Cref{fig:data_processing} (c)) that pass this simulation-based validation are saved as the cleaned dataset ${\mathcal{D}}_{\text{SMPL}}^{\text{Cleaned}}$.

\paragraph{Retargeting SMPL Motions to Robot Motions}

With the cleaned dataset ${\mathcal{D}}_{\text{SMPL}}^{\text{Cleaned}}$ in SMPL format, we retarget the motions into robot motions following the shape-and-motion two-stage retargeting process~\cite{he2024learning}. Since the SMPL parameters estimated by TRAM represent various human body shapes, we first optimize the shape parameter $\boldsymbol{\beta}^{\prime}$ to approximate a humanoid shape. By selecting 12 body links with correspondences between humans and humanoids, we perform gradient descent on $\boldsymbol{\beta}^{\prime}$ to minimize joint distances in the rest pose. Using the optimized shape $\boldsymbol{\beta}^{\prime}$ along with the original translation $\boldsymbol{p}$ and pose $\boldsymbol{\theta}$, we apply gradient descent to further minimize the distances of the body links. This process ensures accurate motion retargeting and produces the cleaned robot trajectory dataset ${\mathcal{D}}_{\text{Robot}}^{\text{Cleaned}}$, as shown in \Cref{fig:data_processing} (d). 

\subsection{Phase-based Motion Tracking Policy Training}

We formulate the motion-tracking problem as a goal-conditioned reinforcement learning (RL) task, where the policy $\pi$ is trained to track the retargeted robot movement trajectories in the dataset ${\mathcal{D}}_{\text{Robot}}^{\text{Cleaned}}$. Inspired by ~\cite{peng2018deepmimic}, the state $s_t$ includes the robot’s proprioception $s_t^{\mathrm{p}}$ and a time phase variable $\phi \in [0,1]$, where $\phi=0$ represents the start of a motion and $\phi=1$ represents the end. This time phase variable alone $\phi$ is proven to be sufficient to serve as the goal state $\boldsymbol{s}_t^{\mathrm{g}}$ for single-motion tracking~\cite{peng2018deepmimic}. 
The proprioception $s_t^{\mathrm{p}}$ is defined as $s_t^{\mathrm{p}} \triangleq \left[\dofposhist, \dofvelhist,  \rootangvelhist, \gravityhist, \actionhist \right]$, with 5-step history of joint position $\dofpos\in\mathbb{R}^{23}$, joint velocity $\dofvel\in\mathbb{R}^{23}$, root angular velocity $\rootangvel\in\mathbb{R}^3$, root projected gravity $\gravity\in\mathbb{R}^3$, and last action $\actionprev\in\mathbb{R}^{23}$.
Using the agent’s proprioception $s_t^{\mathrm{p}}$ and the goal state $\boldsymbol{s}_t^{\mathrm{g}}$, we define the reward as  $r_t=\mathcal{R}\left(s_t^{\mathrm{p}}, s_t^{\mathrm{g}}\right)$, which is used for policy optimization.  The specific reward terms can be found in~\Cref{tab:deepmimic_reward}. The action $\boldsymbol{a}_t \in \mathbb{R}^{23}$ corresponds to the target joint positions and is passed to a PD controller that actuates the robot’s degrees of freedom. To optimize the policy, we use the proximal policy optimization (PPO)~\cite{schulman2017proximal}, aiming to maximize the cumulative discounted reward $\mathbb{E}\left[\sum_{t=1}^T \gamma^{t-1} r_t\right]$. We identify several design choices that are crucial for achieving stable policy training:

\paragraph{Asymmetric Actor-Critic Training}

Real-world humanoid control is inherently a partially observable Markov decision process (POMDP), where certain task-relevant properties that are readily available in simulation become unobservable in real-world scenarios. However, these missing properties can significantly facilitate policy training in simulation. To bridge this gap, we employ an asymmetric actor-critic framework, where the critic network has access to privileged information such as the global positions of the reference motion and the root linear velocity, while the actor network relies solely on proprioceptive inputs and a time-phase variable. This design not only enhances phase-based motion tracking during training but also enables a simple, phase-driven motion goal for sim-to-real transfer. Crucially, because the actor does not depend on position-based motion targets, our approach eliminates the need for odometry during real-world deployment—overcoming a well-documented challenge in prior work on humanoid robots~\cite{he2024learning,he2024omnih2o}.

\paragraph{Termination Curriculum of Tracking Tolerance}
Training a policy to track agile motions in simulation is challenging, as certain motions can be too difficult for the policy to learn effectively. For instance, when imitating a jumping motion, the policy often fails early in training and learns to remain on the ground to avoid landing penalties. To mitigate this issue, we introduce a termination curriculum that progressively refines the motion error tolerance throughout training, guiding the policy toward improved tracking performance. Initially, we set a generous termination threshold of 1.5m, meaning the episode terminates if the robot deviates from the reference motion by this margin. As training progresses, we gradually tighten this threshold to 0.3m, incrementally increasing the tracking demand on the policy. This curriculum allows the policy to first develop basic balancing skills before progressively enforcing stricter motion tracking, ultimately enabling successful execution of high-dynamic behaviors.


\paragraph{Reference State Initialization}
Task initialization plays a crucial role in RL training. We find that naively initializing episodes at the start of the reference motion leads to policy failure. For example, in Cristiano Ronaldo's jumping training, starting the episode from the beginning forces the policy to learn sequentially. However, a successful backflip requires mastering the landing first—if the policy cannot land correctly, it will struggle to complete the full motion from takeoff. To address this, we adopt the Reference State Initialization (RSI) framework~\cite{peng2018deepmimic}. Specifically, we randomly sample time-phase variables between 0 and 1, which effectively randomizes the starting point of the reference motion for the policy to track. We then initialize the robot’s state based on the corresponding reference motion at that phase, including root position and orientation, root linear and angular velocities and joint positions and velocities. This initialization strategy significantly improves motion tracking training, particularly for agile whole-body motions, by allowing the policy to learn different motion phases in parallel rather than being constrained to a strictly sequential learning process.

\paragraph{Reward Terms}
We define the reward function $r_t$ with the sum of three terms: 1) penalty, 2) regularization, and 3) task rewards. A detailed summary of these components is provided in \Cref{tab:deepmimic_reward}.
\begin{table}[!h]
    \centering
    \small 
    \setlength{\tabcolsep}{2pt} 
    \vspace{-5mm}
    \caption{Reward Terms for Pretraining}
    \vspace{-2mm}
    \label{tab:deepmimic_reward}
    \resizebox{0.8\columnwidth}{!}{ 
    \begin{tabular}{cccc}
        \toprule
        Term & Weight & Term & Weight \\
        \midrule
        \multicolumn{4}{c}{Penalty} \\
        \midrule
        DoF position limits & $-10.0$ & DoF velocity limits & $-5.0$ \\
        Torque limits & $-5.0$ & Termination & $-200.0$ \\
        \midrule
        \multicolumn{4}{c}{Regularization} \\
        \midrule
        Torques & $-1 \times 10^{-6}$ & Action rate & $-0.5$ \\
        Feet orientation & $-2.0$ & Feet heading & $-0.1$ \\
        Slippage & $-1.0$ &  \\
        \midrule
        \multicolumn{4}{c}{Task Reward} \\
        \midrule
        Body position & $1.0$ & VR 3-point & $1.6$ \\
        Body position (feet) & $2.1$ & Body rotation & $0.5$ \\
        Body angular velocity & $0.5$ & Body velocity & $0.5$ \\
        DoF position & $0.75$ & DoF velocity & $0.5$ \\
        \bottomrule
    \end{tabular}
    \vspace{-7mm}
    } 
\end{table}

\paragraph{Domain Randomizations}
To improve the robustness of the pre-trained policy in \Cref{fig:ASAP} (a), we utilized basic domain randomization techniques listed in \Cref{tab:deepmimic_DR}.
\section{Post-training: Training Delta Action Model and Fine-tuning Motion Tracking Policy}
The policy trained in the first stage can track the reference motion in the real-world but does not achieve high motion quality. Thus, during the second stage, as shown in ~\Cref{fig:ASAP}~(b) and (c), we leverage real-world data rolled out by the pre-trained policy to train a delta action model, followed by policy refinement through dynamics compensation using this learned delta action model.

\subsection{Data Collection}
\begin{figure}[t]
    \centering
    \includegraphics[width=1\linewidth]{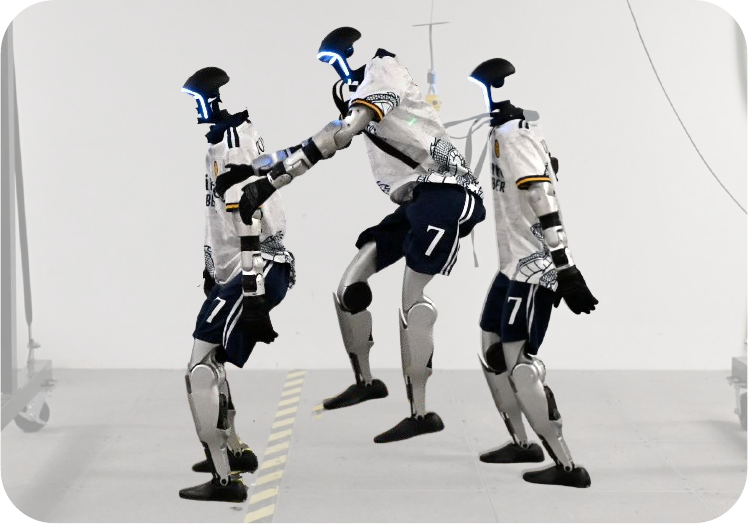}
     \vspace{-3mm}
    \caption{We deploy the pretrained policy of a forward jump motion tracking task, challenging the 1.35m-tall Unitree G1 robot for a forward leap over 1m.}
    \label{fig:data-collect}
    \vspace{-4mm}
\end{figure}

We deploy the pretrained policy in the real world to perform whole-body motion tracking tasks (as depicted in~\Cref{fig:data-collect}) and record the resulting trajectories, denoted as $\mathcal{D}^\text{r} = \{s^\text{r}_0, a^\text{r}_0, \dots, s^\text{r}_T, a^\text{r}_T\}$, as illustrated in~\Cref{fig:ASAP}~(a). At each timestep $t$, we use a motion capture device and onboard sensors to record the state: 
$
s_t = [p^\text{base}_t, v_t^\text{base}, \alpha^\text{base}_t, \omega^\text{base}_t, q_t, \dot{q}_t],
$
where $p^\text{base}_t \in \mathbb{R}^3$ represents the robot base 3D position, $v_t^\text{base} \in \mathbb{R}^3$ is base linear velocity, $\alpha^\text{base}_t \in \mathbb{R}^4$ is the robot base orientation represented as a quaternion, $\omega^\text{base}_t \in \mathbb{R}^3$ is the base angular velocity, $q_t \in \mathbb{R}^{23}$ is the vector of joint positions, and $\dot{q}_t \in \mathbb{R}^{23}$ represents joint velocities.

\begin{figure*}[t]
    \centering
    \includegraphics[width=0.9\textwidth]{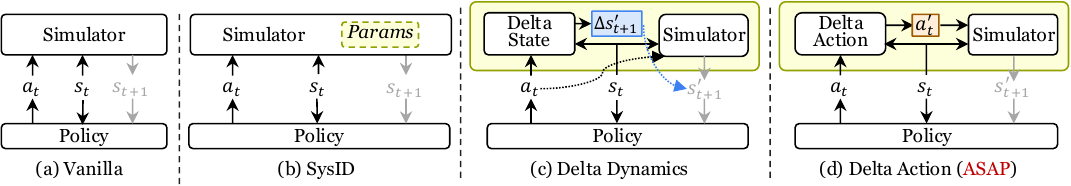}
    \vspace{-1mm}
    \caption{Baselines of \method. (a) Model-free RL training. (b) System ID from real to sim using real-world data. (c) Learning delta dynamics model using real-world data. (d) Our proposed method, learning delta action model using real-world data. }
    \label{fig:baselines}
    \vspace{-4mm}
\end{figure*}

\subsection{Training Delta Action Model}
\label{sec:train-delta-action-model}

Due to the sim-to-real gap, when we replay the real-world trajectories in simulation, the resulting simulated trajectory will likely deviate significantly from real-world recorded trajectories. This discrepancy is a valuable learning signal for learning the mismatch between simulation and real-world physics. We leverage an RL-based delta/residual action model to compensate for the sim-to-real physics gap.

As illustrated in~\Cref{fig:ASAP} (b), the delta action model is defined as $\Delta a_t = \pi^\Delta_\theta(s_t, a_t)$, where the policy $\pi^\Delta_\theta$ learns to output corrective actions based on the current state $s_t$ and the action $a_t$. These corrective actions ($\Delta a_t$) are added to the real-world recorded actions ($a^r_t$) to account for discrepancies between simulation and real-world dynamics.

The RL environment incorporates this delta action model by modifying the simulator dynamics as follows: $ s_{t+1} = f^\text{sim}(s_t, a^r_t + \Delta a_t)$ where $f^\text{sim}$ represents the simulator's dynamics, $a^r_t$ is the reference action recorded from real-world rollouts, and $\Delta a_t$ introduces corrections learned by the delta action model.

\begin{table}[htp]
    \centering
    \vspace{-2mm}
    \small 
    \setlength{\tabcolsep}{2pt} 
    \caption{Reward Terms for Delta Action Learning}
    \vspace{-2mm}
    \label{tab:deltaA_FT_reward}
    \resizebox{0.8\columnwidth}{!}{ 
    \begin{tabular}{cccc}
        \toprule
        Term & Weight & Term & Weight \\
        \midrule
        \multicolumn{4}{c}{Penalty} \\
        \midrule
        DoF position limits & $-10.0$ & DoF velocity limits & $-5.0$ \\
        Torque limits & $-0.1$ & Termination & $-200.0$ \\
        \midrule
        \multicolumn{4}{c}{Regularization} \\
        \midrule
        Action rate & $-0.01$ & Action norm & $-0.2$ \\
        \midrule
        \multicolumn{4}{c}{Task Reward} \\
        \midrule
        Body position & $1.0$ & VR 3-point & $1.0$ \\
        Body position (feet) & $1.0$ & Body rotation & $0.5$ \\
        Body angular velocity & $0.5$ & Body velocity & $0.5$ \\
        DoF position & $0.5$ & DoF velocity & $0.5$ \\
        \bottomrule
    \end{tabular}
    } 
    \vspace{-2mm}
\end{table}

During each RL step: 
\begin{enumerate}
    \item The robot is initialized at the real-world state $s^r_t$.
    \item  A reward signal is computed to minimize the discrepancy between the simulated state $s_{t+1}$ and the recorded real-world state $s^r_{t+1}$, with an additional action magnitude regularization term $\exp(-\lVert a_t \rVert) - 1)$, as specified in~\Cref{tab:deltaA_FT_reward}. The workflow is illustrated in \Cref{fig:ASAP}~(b).
    \item PPO is used to train the delta action policy $\pi^\Delta_\theta$, learning corrected $\Delta a_t$ to match simulation and the real world.
\end{enumerate}

By learning the delta action model, the simulator can accurately reproduce real-world failures. For example, consider a scenario where the simulated robot can jump because its motor strength is overestimated, but the real-world robot cannot jump due to weaker motors. The delta action model $\pi^\Delta_\theta$ will learn to reduce the intensity of lower-body actions, simulating the motor limitations of the real-world robot. This allows the simulator to replicate the real-world dynamics and enables the policy to be fine-tuned to handle these limitations effectively.

\subsection{Fine-tuning Motion Tracking Policy under New Dynamics}
With the learned delta action model $\pi^\Delta (s_t, a_t)$, we can reconstruct the simulation environment with 
$$
s_{t+1} = f^{\text{\method}}(s_t, a_t) = f^\text{sim}(s_t, a_t + \pi^\Delta(s_t, a_t)),
$$
As shown in~\Cref{fig:ASAP} (c), we keep the $\pi^\Delta$ model parameters frozen, and fine-tune the pretrained policy with the same reward summarized in~\Cref{tab:deepmimic_reward}.

\subsection{Policy Deployment}
Finally, we deploy the fine-tuned policy without delta action model in the real world as shown in \Cref{fig:ASAP}~(d). The fine-tuned policy shows enhanced real-world motion tracking performance compared to the pre-trained policy. Quantitative improvements will be discussed in \Cref{sec:EXP1}.

\section{Performance Evaluation of \method}
\label{sec:EXP1}

\begin{figure*}[t]
    \centering
    \includegraphics[width=\textwidth]{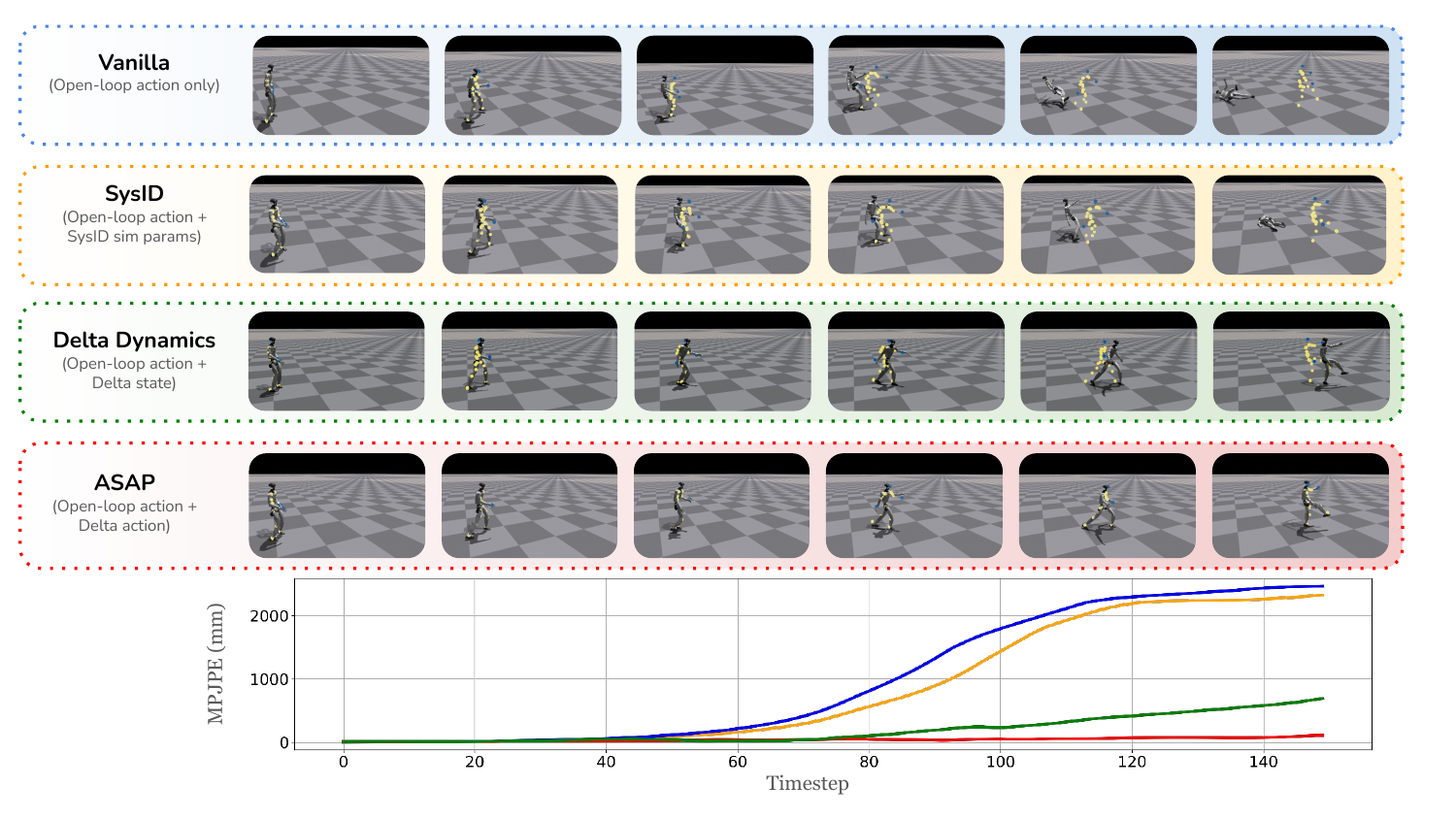}
    \vspace{-6mm}
    \caption{Replaying IsaacSim State-Action trajecories in IsaacGym. The upper four panels visualize the Unitree G1 humanoid executing a soccer-shooting motion under four distinct open-loop actions. Corresponding metric curves (bottom) quantify tracking performance. Importantly, our delta action model (ASAP) is trained across multiple motions and is not overfitted to this specific example.}
    \label{fig:ASAP_openloop_curves}
    \vspace{-3mm}
\end{figure*}

In this section, we present extensive experimental results on three policy transfers: IsaacGym~\cite{makoviychuk2021isaac} to IsaacSim~\cite{mittal2023orbit}, IsaacGym to Genesis~\cite{Genesis}, and IsaacGym to real-world Unitree G1 humanoid robot. Our experiments aim to address the following key questions:
\begin{itemize}
    \item \textbf{Q1}: Can \method outperform other baseline methods to compensate for the dynamics mismatch? 
    \item \textbf{Q2}: Can \method finetune policy to outperform SysID and Delta Dynamics methods? 
    \item \textbf{Q3}: Does \method work for sim-to-real transfer?
\end{itemize}

\textbf{Experiments Setup.}
To address these questions, we evaluate \method on motion tracking tasks in both simulation (\Cref{sec:sim-open-loop} and \Cref{sec:sim-close-loop}) and real-world settings (\Cref{sec:real-exp}).

\begin{figure*}[t]
    \centering
    \includegraphics[width=\linewidth]{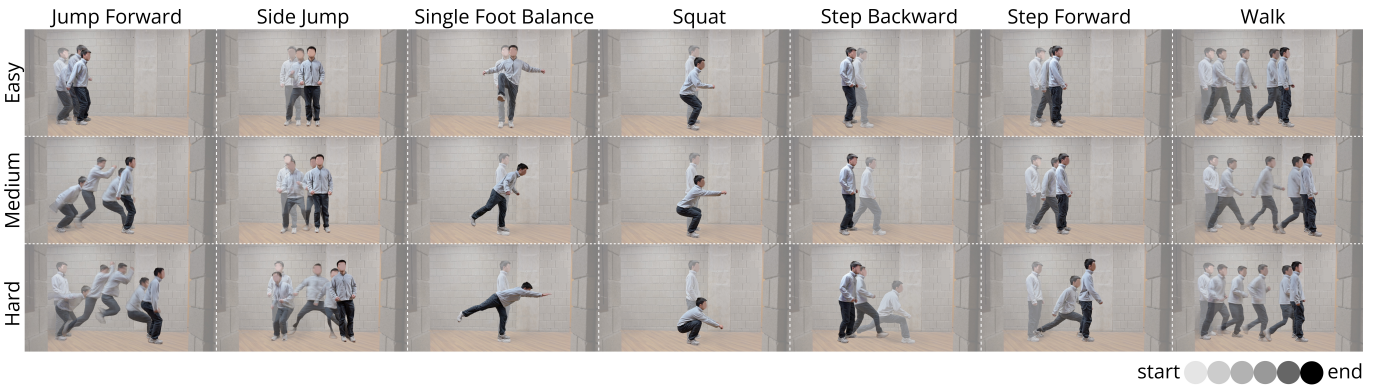}
    \vspace{-6mm}
    \caption{Visual comparisons of motion imitation results across different difficulty levels (Easy, Medium, Hard) for various tasks including Jump Forward, Side Jump, Single Foot Balance, Squat, Step Backward, Step Forward, and Walk.}
    \vspace{-2mm}
    \label{fig:demo_task_difficulty}
\end{figure*}

In the simulation, we use the retargeted motion dataset from the videos we shoot, denoted as ${\mathcal{D}}_{\text{Robot}}^{\text{Cleaned}}$, which contains diverse human motion sequences. We select 43 motions categorized into three difficulty levels: easy, medium, and hard (as partially visualized in~\Cref{fig:demo_task_difficulty}), based on motion complexity and the required agility. \method is evaluated through simulation-to-simulation transfer by training policies in IsaacGym and using two other simulators, IsaacSim and Genesis, as a proxy of ``real-world'' environments. This setup allows for a systematic evaluation of \method's generalization and transferability. The success of the transfer is assessed by metrics described in subsequent sections.

For real-world evaluation, we deploy \method on Unitree G1 robot with fixed wrists to track motion sequences that has obvious sim-to-real gap. These sequences are chosen to capture a broad range of motor capabilities and demonstrate the sim-to-real capability for agile whole-body control.

\textbf{Baselines.} We have the following baselines:

\texttt{Oracle}: This baseline is trained and evaluated entirely within IsaacGym. It assumes perfect alignment between the training and testing environments, serving as an upper bound for performance in simulation.

\texttt{Vanilla} (\Cref{fig:baselines} a): The RL policy is trained in IsaacGym and evaluated in IsaacSim, Genesis, or the real world.

\texttt{SysID} (\Cref{fig:baselines} b): 
We identify the following representative parameters in our simulated model that best align the ones in the real world: base center of mass (CoM) shift $(c_x, c_y, c_z)$, base link mass offset ratio $k_m$ and low-level PD gain ratios $(k^i_p, k^i_d)$ where $i=1,2,...,23$. Specifically, we search the best parameters among certain discrete ranges by replaying the recorded trajectories in real with different simulation parameters summarized in \Cref{tab:sysid_params}. We then finetune the pre-trained policy in IsaacGym with the best SysID parameters.

\texttt{DeltaDynamics} (\Cref{fig:baselines} c): We train a residual dynamics model $f^\Delta_{\theta}(s_t, a_t)$ to capture the discrepancy between simulated and real-world physics. The detailed implementation is introduced in~\Cref{sec:appendix-dynamics}

\textbf{Metrics.}
We report success rate, deeming imitation unsuccessful when, at any point during imitation, the average difference in body distance is on average further than 0.5m.
We evaluate policy’s ability to imitate the reference motion by comparing the tracking error of the global body position $E_\text{g-mpjpe}$ (mm), the root-relative mean per-joint (MPJPE) $E_{\text{mpjpe}}$ (mm), acceleration error $E_\text{acc}$ ($\text{mm/frame}^2$), and root velocity $E_\text{vel}$ (mm/frame). The mean values of the metrics are computed across all motion sequences used.

\begin{table}[t]
\caption{Open-loop performance comparison across simulators and motion lengths.}
\vspace{-2mm}
\label{tab:open-loop}
\centering
\resizebox{\linewidth}{!}{%
\begingroup
\setlength{\tabcolsep}{2pt}
\renewcommand{\arraystretch}{0.8}
\begin{tabular}{llcccccccc}
\toprule
\multicolumn{2}{c}{Simulator \& Length} & \multicolumn{4}{c}{\cellcolor{llnvgreen}IsaacSim} & \multicolumn{4}{c}{\cellcolor{gsred}Genesis} \\
\cmidrule(r){3-6} \cmidrule(r){7-10}
Length & Method &
$E_\text{g-mpjpe}$ & $E_\text{mpjpe}$ & $\text{E}_{\text{acc}}$ & $\text{E}_{\text{vel}}$ & 
$E_\text{g-mpjpe}$ & $E_\text{mpjpe}$ & $\text{E}_{\text{acc}}$ & $\text{E}_{\text{vel}}$ \\
\midrule

\multirow{4}{*}{0.25s} 
& OpenLoop      & \goodnumber{19.5} & 15.1 & \goodnumber{6.44} & \goodnumber{5.80} & 19.8 & 15.3 & 6.53 & 5.88 \\
& SysID         & \goodnumber{19.4} & 15.0 & \goodnumber{6.43} & \goodnumber{5.74} & 19.3 & 15.0 & 6.42 & 5.73 \\
& DeltaDynamics & 24.4 & \goodnumber{13.6} & 9.43 & 7.85 & 20.0 & \goodnumber{12.4} & 8.42 & 6.89 \\
& \method        & \goodnumber{19.9} & 15.6 & \goodnumber{6.48} & \goodnumber{5.86} & \goodnumber{19.0} & 14.9 & \goodnumber{6.19} & \goodnumber{5.59} \\

\midrule
\multirow{4}{*}{0.5s} 
& OpenLoop      & 33.3 & 23.2 & 6.80 & 6.84 & 33.1 & 23.0 & 6.78 & 6.82 \\
& SysID         & 32.1 & 22.2 & 6.57 & 6.56 & 32.2 & 22.3 & 6.57 & 6.57 \\
& DeltaDynamics & 36.5 & \goodnumber{16.4} & 8.89 & 7.98 & 27.8 & \goodnumber{14.0} & 7.63 & 6.74 \\
& \method        & \goodnumber{26.8} & 19.2 & \goodnumber{5.09} & \goodnumber{5.36} & \goodnumber{25.9} & 18.4 & \goodnumber{4.93} & \goodnumber{5.19} \\

\midrule
\multirow{4}{*}{1.0s} 
& OpenLoop      & 80.8 & 43.5 & 10.6 & 11.1 & 82.5 & 44.5 & 10.8 & 11.4 \\
& SysID         & 77.6 & 41.5 & 10.2 & 10.7 & 76.5 & 41.6 & 10.0 & 10.5 \\
& DeltaDynamics & 68.1 & \goodnumber{21.5} & 9.61 & 9.14 & 50.2 & \goodnumber{17.2} & 8.19 & 7.62 \\
& \method        & \goodnumber{37.9} & 22.9 & \goodnumber{4.38} & \goodnumber{5.26} & \goodnumber{36.9} & 22.6 & \goodnumber{4.23} & \goodnumber{5.10} \\

\bottomrule
\end{tabular}
\endgroup
}
\vspace{-3mm}
\end{table}

\begin{table*}[t]
\caption{Closed-loop motion imitation evaluation across different simulators. All variants are trained with identical rewards.}
\label{tab:closed-loop}
\vspace{-2mm}
\centering
\resizebox{\linewidth}{!}{%
\begingroup
\setlength{\tabcolsep}{2pt}
\renewcommand{\arraystretch}{0.8}
\begin{tabular}{llcccccccccc}
\toprule
\multicolumn{2}{c}{Test Environment} & \multicolumn{5}{c}{\cellcolor{llnvgreen}IsaacSim} & \multicolumn{5}{c}{\cellcolor{gsred}Genesis} \\
\cmidrule(r){3-7} \cmidrule(r){8-12}
Level & Method &
$\text{Succ} \uparrow$ & $E_\text{g-mpjpe}  \downarrow$ &  $E_\text{mpjpe} \downarrow $ &  $\text{E}_{\text{acc}} \downarrow$  & $\text{E}_{\text{vel}} \downarrow$ &
$\text{Succ} \uparrow$ & $E_\text{g-mpjpe}  \downarrow$ &  $E_\text{mpjpe} \downarrow $ &  $\text{E}_{\text{acc}} \downarrow$  & $\text{E}_{\text{vel}} \downarrow$ \\
\midrule
& \cellcolor{verylightgray}{Oracle (IsaacGym $\rightarrow$ IsaacGym)} & \cellcolor{verylightgray}{100\%{\tiny\(\pm\text{0.000\%}\)}} & \cellcolor{verylightgray}{97.5{\tiny\(\pm\text{0.605}\)}} & \cellcolor{verylightgray}{43.2{\tiny\(\pm\text{0.112}\)}} & \cellcolor{verylightgray}{2.56{\tiny\(\pm\text{0.024}\)}} & \cellcolor{verylightgray}{4.48{\tiny\(\pm\text{0.023}\)}} & \cellcolor{verylightgray}{100\%{\tiny\(\pm\text{0.000\%}\)}} & \cellcolor{verylightgray}{97.5{\tiny\(\pm\text{0.605}\)}} & \cellcolor{verylightgray}{43.2{\tiny\(\pm\text{0.112}\)}} & \cellcolor{verylightgray}{2.56{\tiny\(\pm\text{0.024}\)}} & \cellcolor{verylightgray}{4.48{\tiny\(\pm\text{0.023}\)}} \\
\cmidrule(lr){2-12}
&  Vanilla (IsaacGym $\rightarrow$ TestEnv) & 100\%{\tiny\(\pm\text{0.000\%}\)} & \goodnumber{107} {\tiny\(\pm\text{0.578}\)} & 45.4{\tiny\(\pm\text{0.169}\)} & 2.83{\tiny\(\pm\text{0.012}\)} &  4.59{\tiny\(\pm\text{0.021}\)} &
100\%{\tiny\(\pm\text{0.000\%}\)} & 140{\tiny\(\pm\text{1.85}\)} & \goodnumber{70.1}{\tiny\(\pm\text{0.626}\)} & 2.68{\tiny\(\pm\text{0.042}\)} & 4.65{\tiny\(\pm\text{0.046}\)} \\
\multirow{-0.7}{*}{Easy} & 
SysID & 100\%{\tiny\(\pm\text{0.000\%}\)} & \goodnumber{105} {\tiny\(\pm\text{1.35}\)} & 47.8{\tiny\(\pm\text{0.970}\)} & 3.09{\tiny\(\pm\text{0.011}\)} & 4.98{\tiny\(\pm\text{0.020}\)} & 
100\%{\tiny\(\pm\text{0.000\%}\)} & \goodnumber{127}{\tiny\(\pm\text{0.233}\)} & 79.9{\tiny\(\pm\text{0.330}\)} & 2.99{\tiny\(\pm\text{0.035}\)} & 4.95{\tiny\(\pm\text{0.012}\)} \\
& DeltaDynamics & 100\%{\tiny\(\pm\text{0.000\%}\)} & 127{\tiny\(\pm\text{2.97}\)} & 56.7{\tiny\(\pm\text{0.390}\)} &  3.50{\tiny\(\pm\text{0.028}\)} & 5.56{\tiny\(\pm\text{0.031}\)} & 83.3\%{\tiny\(\pm\text{0.000\%}\)} & 168{\tiny\(\pm\text{7.62}\)} & 87.0{\tiny\(\pm\text{1.51}\)} & 3.08{\tiny\(\pm\text{0.18}\)} & 5.39{\tiny\(\pm\text{0.34}\)} \\
& \method & 100\%{\tiny\(\pm\text{0.000\%}\)} & \goodnumber{106} {\tiny\(\pm\text{0.498}\)} & \goodnumber{44.3} {\tiny\(\pm\text{0.103}\)} & \goodnumber{2.74} {\tiny\(\pm\text{0.025}\)} & \goodnumber{4.46} {\tiny\(\pm\text{0.020}\)} & 
100\%{\tiny\(\pm\text{0.000\%}\)} & \goodnumber{125}{\tiny\(\pm\text{4.75}\)} & 73.5{\tiny\(\pm\text{0.570}\)} & \goodnumber{2.10}{\tiny\(\pm\text{0.083}\)} & \goodnumber{4.11}{\tiny\(\pm\text{0.133}\)} \\

\midrule
& \cellcolor{verylightgray}{Oracle (IsaacGym $\rightarrow$ IsaacGym)} & \cellcolor{verylightgray}{100\%{\tiny\(\pm\text{0.000\%}\)}} & \cellcolor{verylightgray}{111{\tiny\(\pm\text{0.635}\)}} & \cellcolor{verylightgray}{48.8{\tiny\(\pm\text{0.133}\)}}  & \cellcolor{verylightgray}{2.63{\tiny\(\pm\text{0.017}\)}}  & \cellcolor{verylightgray}{4.82{\tiny\(\pm\text{0.019}\)}} & \cellcolor{verylightgray}{100\%{\tiny\(\pm\text{0.000\%}\)}} & \cellcolor{verylightgray}{111{\tiny\(\pm\text{0.635}\)}} & \cellcolor{verylightgray}{48.8{\tiny\(\pm\text{0.133}\)}}  & \cellcolor{verylightgray}{2.63{\tiny\(\pm\text{0.017}\)}}  & \cellcolor{verylightgray}{4.82{\tiny\(\pm\text{0.019}\)}}  \\
\cmidrule(lr){2-12}
&  Vanilla (IsaacGym $\rightarrow$ TestEnv) & 
100\%{\tiny\(\pm\text{0.000\%}\)} & \goodnumber{114}{\tiny\(\pm\text{0.720}\)} & \goodnumber{49.2}{\tiny\(\pm\text{0.104}\)} & 2.92{\tiny\(\pm\text{0.021}\)} &  5.07{\tiny\(\pm\text{0.016}\)} & 
94.3\%{\tiny\(\pm\text{7.00\%}\)} & 169{\tiny\(\pm\text{5.76}\)} & 72.0{\tiny\(\pm\text{0.692}\)} & 3.26{\tiny\(\pm\text{0.076}\)} & 5.86{\tiny\(\pm\text{0.101}\)} \\
\multirow{-0.7}{*}{Medium} & 
SysID & 100\%{\tiny\(\pm\text{0.000\%}\)} & \goodnumber{115}{\tiny\(\pm\text{1.256}\)} & \goodnumber{49.1}{\tiny\(\pm\text{0.560}\)} & 3.43{\tiny\(\pm\text{0.021}\)} & 5.01{\tiny\(\pm\text{0.017}\)} & 
100\%{\tiny\(\pm\text{0.000\%}\)} & 138{\tiny\(\pm\text{2.70}\)} & 75.4{\tiny\(\pm\text{1.18}\)} & 3.14{\tiny\(\pm\text{0.042}\)} & 5.50{\tiny\(\pm\text{0.058}\)} \\
& DeltaDynamics & 83.3\%{\tiny\(\pm\text{0.000\%}\)} & 151{\tiny\(\pm\text{2.62}\)} & 68.0{\tiny\(\pm\text{0.364}\)} & 2.90{\tiny\(\pm\text{0.047}\)} & 5.90{\tiny\(\pm\text{0.107}\)} & 83.3\%{\tiny\(\pm\text{0.000\%}\)} & 190{\tiny\(\pm\text{1.46}\)} & 89.4{\tiny\(\pm\text{0.50}\)} & 3.44{\tiny\(\pm\text{0.16}\)} & 7.49{\tiny\(\pm\text{0.11}\)} \\
& \method & 100\%{\tiny\(\pm\text{0.000\%}\)} & \goodnumber{112} {\tiny\(\pm\text{1.648}\)} & \goodnumber{49.3} {\tiny\(\pm\text{0.574}\)} & \goodnumber{2.53} {\tiny\(\pm\text{0.019}\)} & \goodnumber{4.45} {\tiny\(\pm\text{0.026}\)} & 
100\%{\tiny\(\pm\text{0.000\%}\)} & \goodnumber{126}{\tiny\(\pm\text{1.63}\)} & \goodnumber{71.2}{\tiny\(\pm\text{0.163}\)} & \goodnumber{2.81}{\tiny\(\pm\text{0.037}\)} & \goodnumber{5.13}{\tiny\(\pm\text{0.066}\)} \\
\midrule
& \cellcolor{verylightgray}{Oracle (IsaacGym $\rightarrow$ IsaacGym)} & 1\cellcolor{verylightgray}{00\%{\tiny\(\pm\text{0.000\%}\)}} & \cellcolor{verylightgray}{116{\tiny\(\pm\text{0.711}\)}} & \cellcolor{verylightgray}{52.5{\tiny\(\pm\text{0.298}\)}} & \cellcolor{verylightgray}{3.40{\tiny\(\pm\text{0.027}\)}} & \cellcolor{verylightgray}{6.16 {\tiny\(\pm\text{0.028}\)}} & \cellcolor{verylightgray}{100\%{\tiny\(\pm\text{0.000\%}\)}} & \cellcolor{verylightgray}{116{\tiny\(\pm\text{0.711}\)}} & \cellcolor{verylightgray}{52.5{\tiny\(\pm\text{0.298}\)}} & \cellcolor{verylightgray}{3.40{\tiny\(\pm\text{0.027}\)}} & \cellcolor{verylightgray}{6.16 {\tiny\(\pm\text{0.028}\)}} \\
\cmidrule(lr){2-12}
&  Vanilla (IsaacGym $\rightarrow$ TestEnv) & 
100\%{\tiny\(\pm\text{0.000\%}\)} & 148{\tiny\(\pm\text{0.845}\)} & \goodnumber{51.6}{\tiny\(\pm\text{0.137}\)} &  4.41{\tiny\(\pm\text{0.055}\)} & 6.88{\tiny\(\pm\text{0.064}\)} & 
82.9\%{\tiny\(\pm\text{5.70\%}\)} & 175{\tiny\(\pm\text{9.77}\)} & 80.7{\tiny\(\pm\text{1.69}\)} & 3.87{\tiny\(\pm\text{0.175}\)} & 7.19{\tiny\(\pm\text{0.199}\)} \\
\multirow{-0.7}{*}{Hard} & 
SysID & 100\%{\tiny\(\pm\text{0.000\%}\)} & 165{\tiny\(\pm\text{3.83}\)} & 58.4{\tiny\(\pm\text{0.229}\)} & 4.87{\tiny\(\pm\text{0.197}\)} & 7.13{\tiny\(\pm\text{0.131}\)} & 
100\%{\tiny\(\pm\text{0.000\%}\)} & 186{\tiny\(\pm\text{3.84}\)} & 93.0{\tiny\(\pm\text{1.49}\)} & 4.98{\tiny\(\pm\text{0.245}\)} & 8.98{\tiny\(\pm\text{0.119}\)} \\
& DeltaDynamics & 66.7\%{\tiny\(\pm\text{0.000\%}\)} & 137{\tiny\(\pm\text{2.59}\)} & 60.2{\tiny\(\pm\text{0.477}\)} &  4.20{\tiny\(\pm\text{0.041}\)} & 7.10{\tiny\(\pm\text{0.024}\)} & 60.0\%{\tiny\(\pm\text{5.70\%}\)} & 190{\tiny\(\pm\text{14.0}\)} & 89.6{\tiny\(\pm\text{9.34}\)} & 4.29{\tiny\(\pm\text{1.16}\)} & 8.70{\tiny\(\pm\text{2.33}\)} \\
& \method & 100\%{\tiny\(\pm\text{0.000\%}\)} & \goodnumber{129} {\tiny\(\pm\text{1.57}\)} & 56.5 {\tiny\(\pm\text{1.15}\)} & \goodnumber{3.72} {\tiny\(\pm\text{0.036}\)} & \goodnumber{6.52} {\tiny\(\pm\text{0.042}\)} & 
100\%{\tiny\(\pm\text{0.000\%}\)} & \goodnumber{129}{\tiny\(\pm\text{2.31}\)} & \goodnumber{77.0}{\tiny\(\pm\text{1.07}\)} & \goodnumber{2.69}{\tiny\(\pm\text{0.040}\)} & \goodnumber{5.65}{\tiny\(\pm\text{0.073}\)} \\
\bottomrule
\end{tabular}
\endgroup}
\end{table*}

\subsection{Comparison of Dynamics Matching Capability}
\label{sec:sim-open-loop}

To address \textbf{Q1} (\textit{Can \method outperform other baseline methods to compensate for the dynamics mismatch?}), we establish sim-to-sim transfer benchmarks to assess the effectiveness of different methods in bridging the dynamics gap. IsaacGym serves as the \textit{training environment}, while IsaacSim and Genesis function as \textit{testing environments}. The primary objective is to evaluate the generalization capability of each approach when exposed to new dynamics conditions.
\textit{Open-loop} evaluation measures how accurately a method can reproduce testing-environment trajectories in the training environment. This is achieved by rolling out the same trajectory executed in the testing environment and assessing tracking discrepancies using key metrics such as MPJPE. An ideal method should minimize the discrepancies between training and testing trajectories when replaying testing-environment actions, thereby demonstrating an improved capacity for compensating dynamics mismatch. Quantitative results in~\Cref{tab:open-loop} demonstrate that \method consistently outperforms the OpenLoop baseline across all replayed motion lengths, achieving lower $E_\text{g-mpjpe}$ and $E_\text{mpjpe}$ values, which indicate improved alignment with testing-env trajectories. 
While SysID helps address short-horizon dynamics gaps, it struggles with long-horizon scenarios due to cumulative error buildup. DeltaDynamics improves upon both SysID and OpenLoop for long horizons but suffers from overfitting, as evidenced by cascading errors magnified over time, as shown in~\Cref{fig:ASAP_openloop_curves}. 
\method, however, demonstrates superior generalization by learning residual policies that effectively bridge the dynamics gap. Comparable trends are observed in the Genesis simulator, where \method achieves notable improvements across all metrics relative to the baseline. These results emphasize the efficacy of learning delta action model to reduce the physics gap and improve open-loop replay performance.

\begin{figure*}[t]
    \centering
    \includegraphics[width=0.9\textwidth]{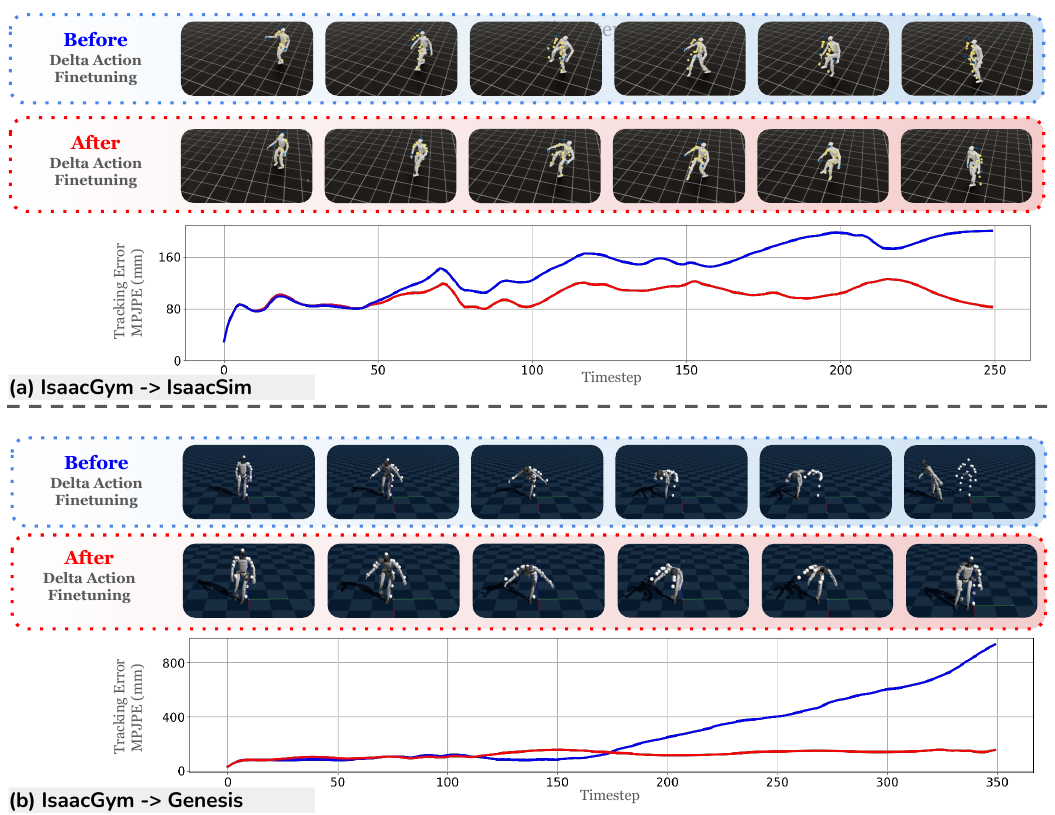}    \caption{Visualization of G1 motion tracking before and after \method fine-tuning in IsaacGym, IsaacSim and Genesis. Top: LeBron James’ “Silencer” motion tracking policy fine-tuning for IsaacGym to IsaacSim. Bottom: \textit{single foot balance} motion tracking policy fine-tuning for IsaacGym to Genesis.}
    \vspace{-2mm}
    \label{fig:ASAP_sim_close_loop}
\end{figure*}

\subsection{Comparison of Policy Fine-Tuning Performance}
\label{sec:sim-close-loop}
To address \textbf{Q2} (\textit{Can \method finetune policy to outperform SysID and Delta Dynamics methods?}), we evaluate the effectiveness of different methods in fine-tuning RL policies for improved testing-environment performance. 
We fine-tune RL policies in modified training environments and subsequently deploy them in the testing environments, quantifying motion-tracking errors in testing environments.
As shown in~\Cref{tab:closed-loop}, \method consistently outperforms baselines such as Vanilla, SysID, and DeltaDynamics across all difficulty levels (Easy, Medium, and Hard) in both simulators (IsaacSim and Genesis). For the Easy level, our method achieves the lowest $E_\text{g-mpjpe}$ and $E_\text{mpjpe}$ in IsaacSim ($E_\text{g-mpjpe} = 106$ and $E_\text{mpjpe} = 44.3$) and Genesis ($E_\text{g-mpjpe} = 125$ and $E_\text{mpjpe} = 73.5$), with minimal acceleration ($\text{E}_{\text{acc}}$) and velocity ($\text{E}_{\text{vel}}$) errors. In more challenging tasks, such as the Hard level, our method continues to excel, significantly reducing motion-tracking errors. For instance, in Genesis, it achieves $E_\text{g-mpjpe} = 129$ and $E_\text{mpjpe} = 77.0$, outperforming SysID and DeltaDynamics by substantial margins. Additionally, our method consistently maintains a 100\% success rate across both simulators, unlike DeltaDynamics, which experiences lower success rates in harder environments. 
To further illustrate the advantages of \method, we provide per-step visualizations in \Cref{fig:ASAP_sim_close_loop}, comparing \method with RL policies deployed without fine-tuning. These visualizations demonstrate that \method successfully adapts to new dynamics and maintains stable tracking performance, whereas baseline methods accumulate errors over time, leading to degraded tracking capability.
These results highlight the robustness and adaptability of our approach in addressing the sim-to-real gap while preventing overfitting and exploitation. The findings validate that \method is an effective paradigm for improving closed-loop performance and ensuring reliable deployment in complex real-world scenarios.

\subsection{Real-World Evaluations}
\label{sec:real-exp}
To answer \textbf{Q3} (\textit{Does \method work for sim-to-real transfer?}). We validate \method on real-world Unitree G1 robot.

\textbf{Real-World Data.}
\label{subsec:real-world-data}In the real-world experiments, we prioritize both motion safety and representativeness by selecting five motion-tracking tasks, including (i)~\textit{kick}, (ii)~\textit{jump forward}, (iii)~\textit{step forward and back}, (iv)~\textit{single foot balance} and (v)~\textit{single foot jump}. 
However, collecting over 400 real-world motion clips— the minimum required to train the full 23-DoF delta action model in simulation, as discussed in\Cref{sec:train-delta-action-model}—poses significant challenges. Our experiments involve highly dynamic motions that cause rapid overheating of joint motors, leading to hardware failures (two Unitree G1 robots broke during data collection). 
Given these constraints, we adopt a more sample-efficient approach by focusing exclusively on learning a 4-DoF ankle delta action model rather than the full-body 23-DoF model. This decision is motivated by two key factors: (1) the limited availability of real-world data makes training the full 23-DoF delta action model infeasible, and (2) the Unitree G1 robot~\cite{Unitree2024G1} features a mechanical linkage design in the ankle, which introduces a significant sim-to-real gap that is difficult to bridge with conventional modeling techniques~\cite{KimJLALS20}.
Under this setting, the original 23 DoF delta action model reduces to 4 DoF delta action model, which needs much less data to be trainable. In practice, we collect 100 motion clips, which prove sufficient to train an effective 4-DoF delta action model for real-world scenarios.

We execute the tracking policy 30 times for each task. In addition to these motion-tracking tasks, we also collect 10 minutes of locomotion data. The locomotion policy will be addressed in the next section, which is also utilized to bridge different tracking policies.

\begin{figure*}[t]
    \centering
    \includegraphics[width=1.0\textwidth]{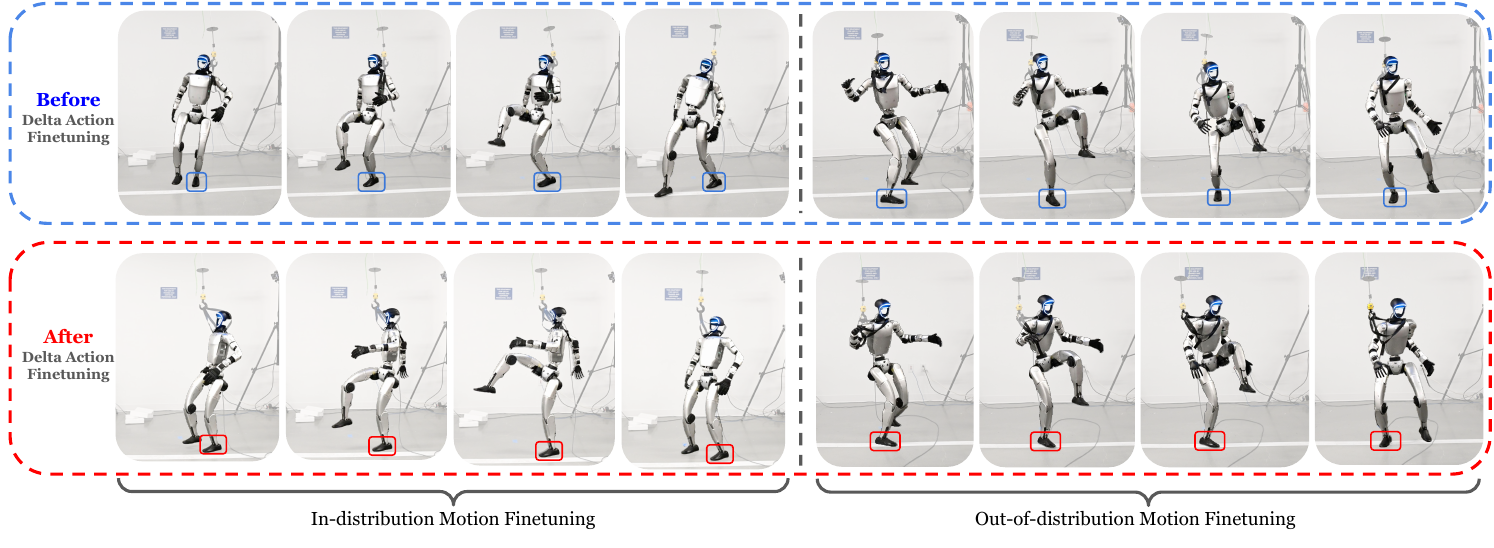}
    \caption{Visualization of LeBron James’ ``Silencer'' motion on the G1 robot before (upper figure enclosed in \textcolor{blue}{blue}) and after (bottom figure enclosed in \textcolor{red}{red}) \method policy finetuning. The left half shows the policy finetuning for the in-distribution motions while the right half shows the out-of-distribution ones. After \method finetuning, the robot behaves more smoothly and reduces jerky lower-body motions.}
    \label{fig:ASAP_real_compare}

\end{figure*}

\textbf{Policy Transition.}
In the real world, we cannot easily reset the robot as in simulators, and therefore we train a robust locomotion policy for the policy transition between different motion-tracking tasks. Our locomotion command contains $(v, \omega, \Pi)$, where $v$ and $\omega$ indicate the linear and angular velocities while $\Pi$ indicates the command to walk or stand still. After each motion-tracking task is done, our locomotion policy will take over to keep the robot balance until the next motion-tracking task begins. In this way, the robot is able to execute multiple tasks without manually resetting.

\textbf{Real-World Results.}
The \simtoreal gap is more pronounced than simulator-to-simulator discrepancies due to factors such as noisy sensor input, inaccuracies in robot modeling, and actuator differences.
To evaluate the effectiveness of \method in addressing these gaps, we compare the closed-loop performance of \method with the \textit{Vanilla} baseline on two representative motion tracking tasks  (kicking and ``Silencer'') in which observe obvious \simtoreal gaps. 
To show the generalizability of the learned delta action model for out-of-distribution motions, we also fine-tune the policy for LeBron James’ ``Silencer'' motion as shown in~\Cref{fig:firstpage} and~\Cref{fig:ASAP_real_compare}. The experiment data is summarized in \Cref{tab:real-exp-close-loop}. It shows that \method outperforms the baseline on both in-distribution and out-of-distribution humanoid motion tracking tasks, achieving a considerable reduction of the tracking errors across all key metrics ($E_\text{g-mpjpe}, E_\text{mpjpe}, E_{acc}$ and $E_{vel}$). These findings highlight the effectiveness of \method in improving \simtoreal transfer for agile humanoid motion tracking.

\begin{table}[t]
\caption{Real-world closed-loop performance comparing with and without \method finetuning on one in-distribution motion and one out-of-distribution motion.}
\label{tab:real-exp-close-loop}
\vspace{-2mm}
\centering
\resizebox{\linewidth}{!}{%
\begingroup
\setlength{\tabcolsep}{2pt}
\renewcommand{\arraystretch}{0.8}
\begin{tabular}{lcccccccc}
\toprule
\multicolumn{1}{c}{Motion} & \multicolumn{4}{c}{\cellcolor{rqblue}Real-World-Kick} & \multicolumn{4}{c}{\cellcolor{rqblue}Real-World-LeBron (OOD)} \\
\cmidrule(r){2-5} \cmidrule(r){6-9}
Method &
$E_\text{g-mpjpe}$ & $E_\text{mpjpe}$ & $\text{E}_{\text{acc}}$ & $\text{E}_{\text{vel}}$ & 
$E_\text{g-mpjpe}$ & $E_\text{mpjpe}$ & $\text{E}_{\text{acc}}$ & $\text{E}_{\text{vel}}$ \\
\midrule

Vanilla      & 61.2 & 43.5 & 2.96 & 2.91 & 159 & 55.3 & 3.43 & 6.43 \\
\method        & \goodnumber{50.2} & \goodnumber{40.1} & \goodnumber{2.46} & \goodnumber{2.70} & \goodnumber{112} & \goodnumber{47.5} & \goodnumber{2.84} & \goodnumber{5.94} \\
\bottomrule
\end{tabular}
\endgroup
}
\end{table}

\section{EXTENSIVE STUDIES AND ANALYSES}

In this section, we aim to thoroughly analyze \method by addressing three central research questions:
\begin{itemize}
    \item \textbf{Q4}: How to best train the delta action model of \method?
    \item \textbf{Q5}: How to best use the delta action model of \method?
    \item \textbf{Q6}: Why and how does \method work?
\end{itemize}

\begin{figure*}[t]
    \centering
    \includegraphics[width=1.0\linewidth]{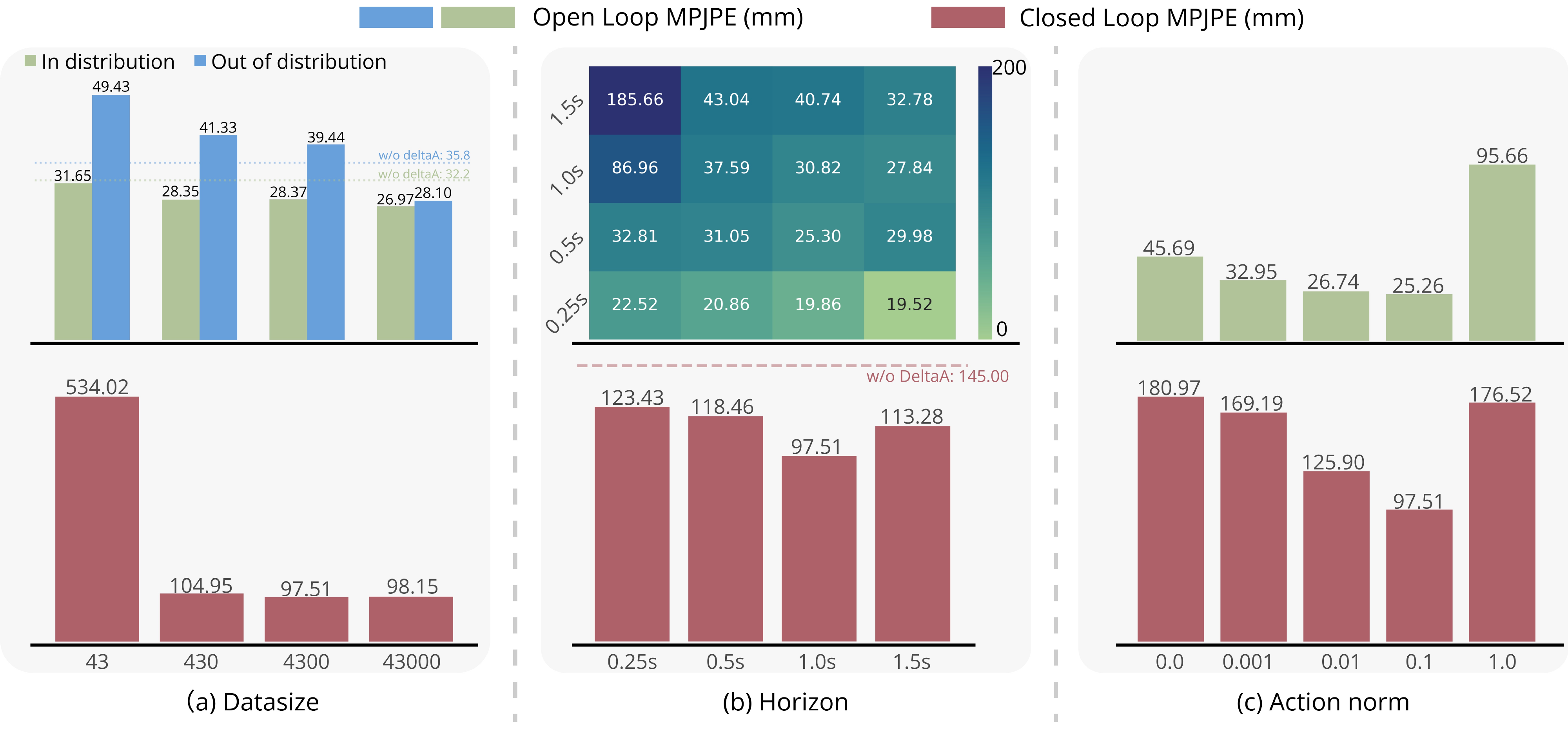}
     \vspace{-4mm}
    \caption{Analysis of dataset size, training horizon, and action norm on the performance of $\pi^\Delta$. (a) \textbf{Dataset Size}: Mean Per Joint Position Error (MPJPE) is evaluated for both in-distribution (green) and out-of-distribution (blue) scenarios. Increasing dataset size leads to enhanced generalization, evidenced by decreasing errors in out-of-distribution evaluations. Closed-loop MPJPE (red bars) also shows improvement with larger datasets. (b) \textbf{Training Horizon}: Open-loop MPJPE (heatmap) improves across evaluation points as training horizons increase, achieving the lowest error at 1.5s. However, closed-loop MPJPE (red bars) shows a sweet spot at a training horizon of 1.0s, beyond which no further improvements are observed. The red dashed line represents the pretrained baseline without $\pi^\Delta$ fine-tuning. (c) \textbf{Action Norm}: The action norm weight significantly influences performance. Both open-loop and closed-loop MPJPE decrease as the weight increases up to 0.1, achieving the lowest error. However, further increases in the action norm weight result in degradation of open-loop performance, highlighting the trade-off between action smoothness and policy flexibility.}
    \label{fig:deltaA_ablation}
\end{figure*}

\subsection{Key Factors in Training Delta Action Models}
\label{sec:VA}
To Answer \textbf{Q4} (\textit{How to best train the delta action model of \method}). 
we conduct a systematic study on key factors influencing the performance of the delta action model. 
Specifically, we investigate the impact of dataset size, training horizon, and action norm weight, evaluating their effects on both open-loop and closed-loop performance. Our analysis uncovers the essential principles for effectively training a high-performing delta action model.

\paragraph{Dataset Size} We analyze the impact of dataset size on the training and generalization of $\pi^\Delta$. Simulation data is collected in Isaac Sim, and $\pi^\Delta$ is trained in Isaac Gym. Open-loop performance is assessed on both in-distribution (training) and out-of-distribution (unseen) trajectories, while closed-loop performance is evaluated using the fine-tuned policy in Isaac Sim. As shown in~\Cref{fig:deltaA_ablation}~(a), increasing the dataset size improves $\pi^\Delta$’s generalization, evidenced by reduced errors in out-of-distribution evaluations. However, the improvement in closed-loop performance saturates, with a marginal decrease of only $0.65\%$ when scaling from $4300$ to $43000$ samples, suggesting limited additional benefit from larger datasets.

\paragraph{Training Horizon} The rollout horizon plays a crucial role in learning $\pi^\Delta$. As shown in~\Cref{fig:deltaA_ablation}~(b), longer training horizons generally improve open-loop performance, with a horizon of 1.5s achieving the lowest errors across evaluation points at 0.25s, 0.5s, and 1.0s. However, this trend does not consistently extend to closed-loop performance. The best closed-loop results are observed at a training horizon of 1.0s, indicating that excessively long horizons do not provide additional benefits for fine-tuned policy.

\paragraph{Action Norm Weight} Training $\pi^\Delta$ incorporates an action norm reward to balance dynamics alignment and minimal correction. As illustrated in~\Cref{fig:deltaA_ablation}~(c), both open-loop and closed-loop errors decrease as the action norm weight increases, reaching the lowest error at a weight of $0.1$. However, further increasing the action norm weight causes open-loop errors to rise, likely due to the minimal action norm reward dominates in the delta action RL training. This highlights the importance of carefully tuning the action norm weight to achieve optimal performance.

\subsection{Different Usage of Delta Action Model}
To answer \textbf{Q5} \textit(How to best use the delta action model of \method?), we compare multiple strategies: fixed-point iteration, gradient-based optimization, and reinforcement learning (RL). Given a learned delta policy \(\pi^\Delta\) such that:
\[
f^\text{sim}(s, a + \pi^\Delta(s, a)) \approx f^\text{real}(s, a),
\]
and a nominal policy \(\hat{\pi}(s)\) that performs well in simulation, the goal is to fine-tune \(\hat{\pi}(s)\) for real-world deployment.

A simple approach is one-step dynamics matching, which leads to the relationship:
\[
\pi(s) = \hat{\pi}(s) - \pi^\Delta(s, \pi(s)).
\]
We consider two RL-free methods: fixed-point iteration and gradient-based optimization. Fixed-point iteration refines \(\hat\pi(s)\) iteratively, while gradient-based optimization minimizes a loss function to achieve a better estimate. These methods are compared against RL fine-tuning, which adapts \(\hat\pi(s)\) using reinforcement learning in simulation. The detailed derivation of these two baselines is summarized in \Cref{sec:appendix_more_deltaA_usage}.

\begin{figure}[htp]
    \centering
    \includegraphics[width=0.85\linewidth]{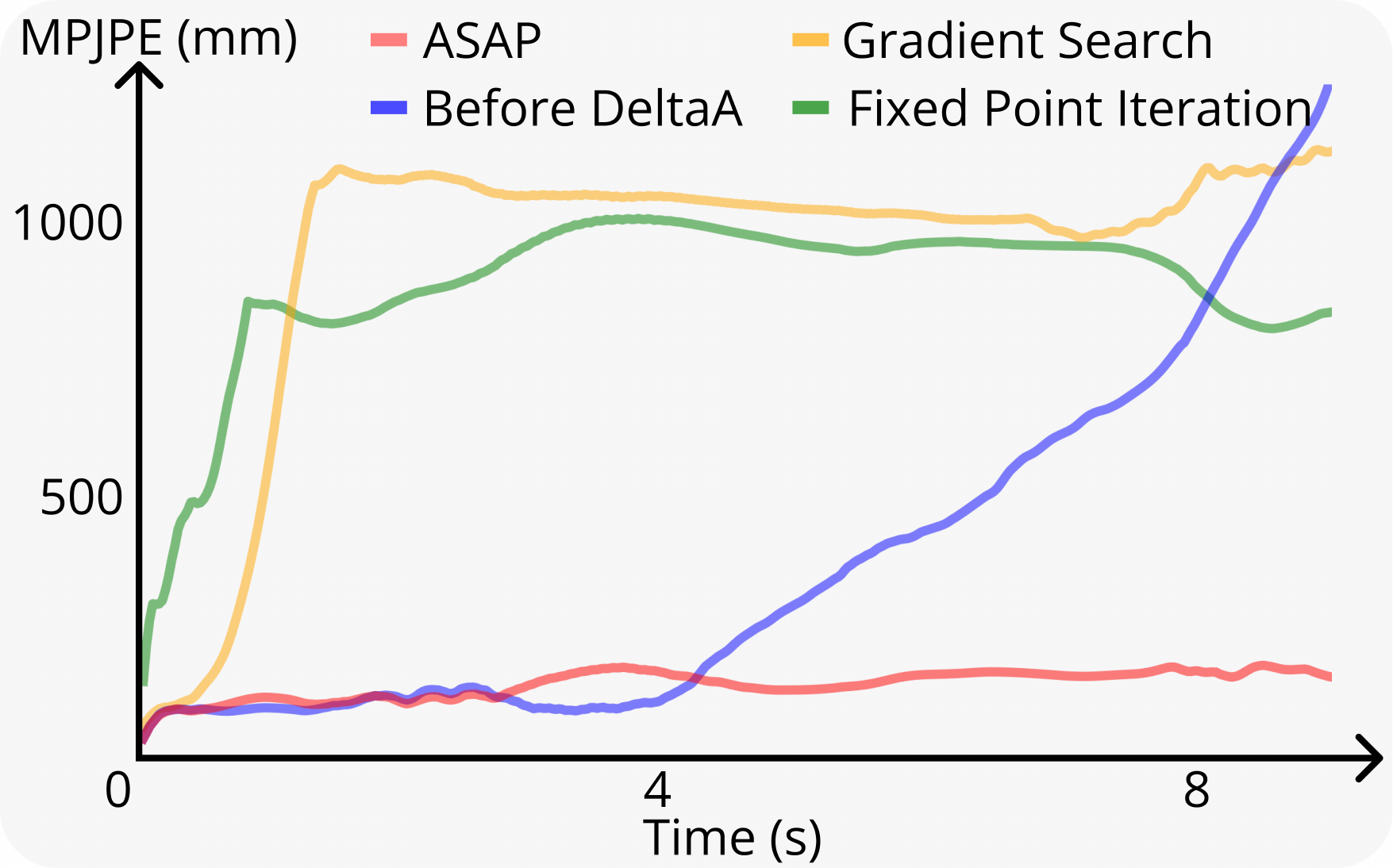}
    \vspace{-2mm}
    \caption{MPJPE comparison over timesteps for fine-tuning methods using delta actionmodel. RL Fine-Tuning achieves the lowest error, while Fixed-Point Iteration and Gradient Search perform worse than the baseline (Before DeltaA) showing the highest error.}
    \label{fig: use_deltaA}
    \vspace{-4mm}
\end{figure}

Our experiments in \Cref{fig: use_deltaA} show that RL fine-tuning achieves the lowest tracking error during deployment, outperforming training-free methods. 
Both RL-free approaches are myopic and suffer from out-of-distribution issues, limiting their real-world applicability (more discussions in \Cref{sec:appendix_more_deltaA_usage}).

\begin{figure}[t]
    \centering
    \includegraphics[width=0.9\linewidth]{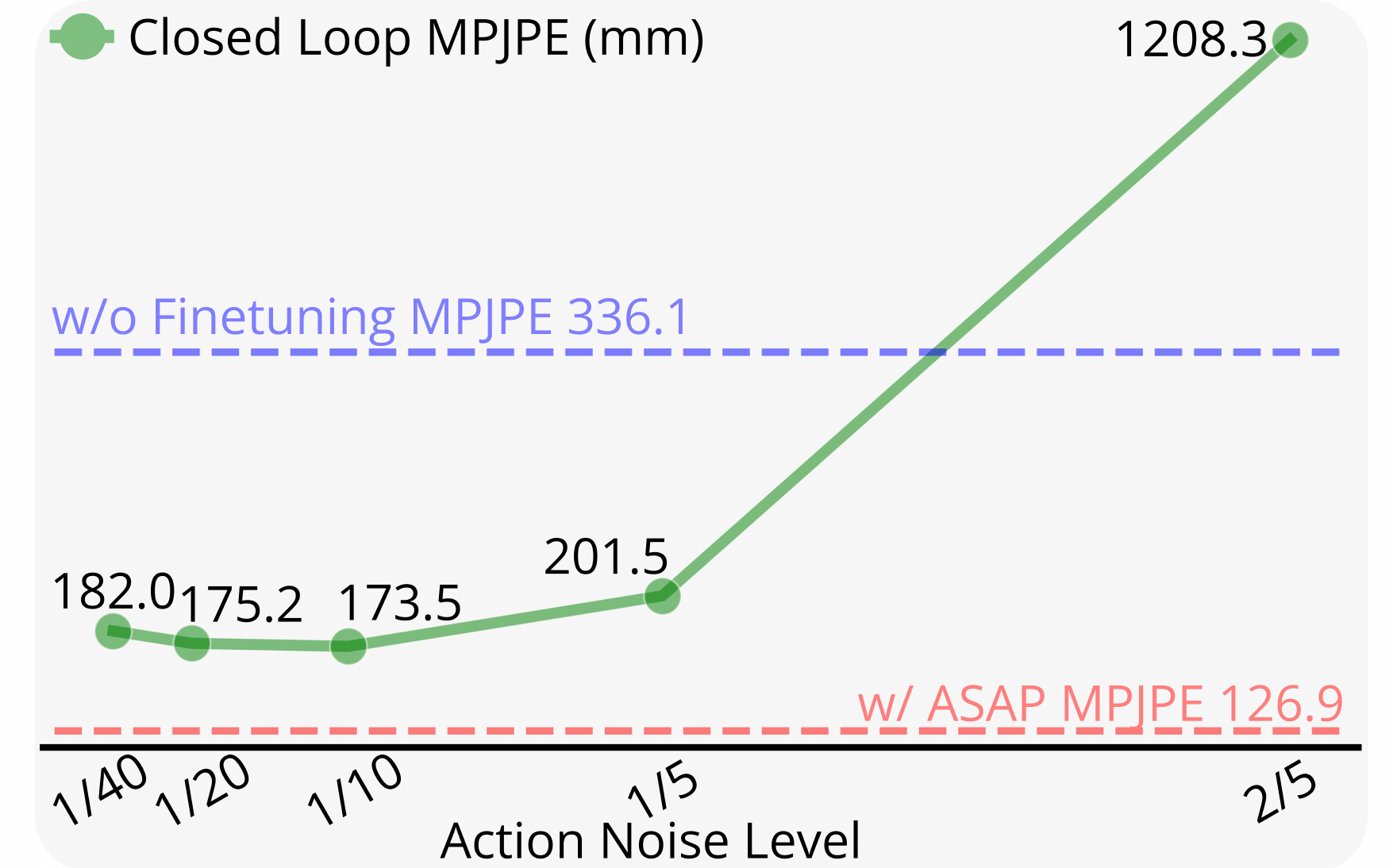}
    \vspace{-2mm}
    \caption{MPJPE vs. Noise Level for policies fine-tuned with random action noise. Policies with noise levels $\beta \in [0.025, 0.2]$ show improved performance compared to no fine-tuning. Delta action achieves better tracking precision (126 MPJPE) compared to the best action noise (173 MPJPE).}
    \label{fig:action_noise}
    \vspace{-4mm}
\end{figure}

\subsection{Does \method Fine-Tuning Outperform Random Action Noise Fine-Tuning?}
To answer \textbf{Q6} (\textit{How does \method work?}), we validate \method finetuning is better than injecting random-action-noise-based finetuning. And we visualize the average magnitude of the delta action model for each joint.

\begin{figure}[t]
    \centering
    \includegraphics[width=1.0\linewidth]{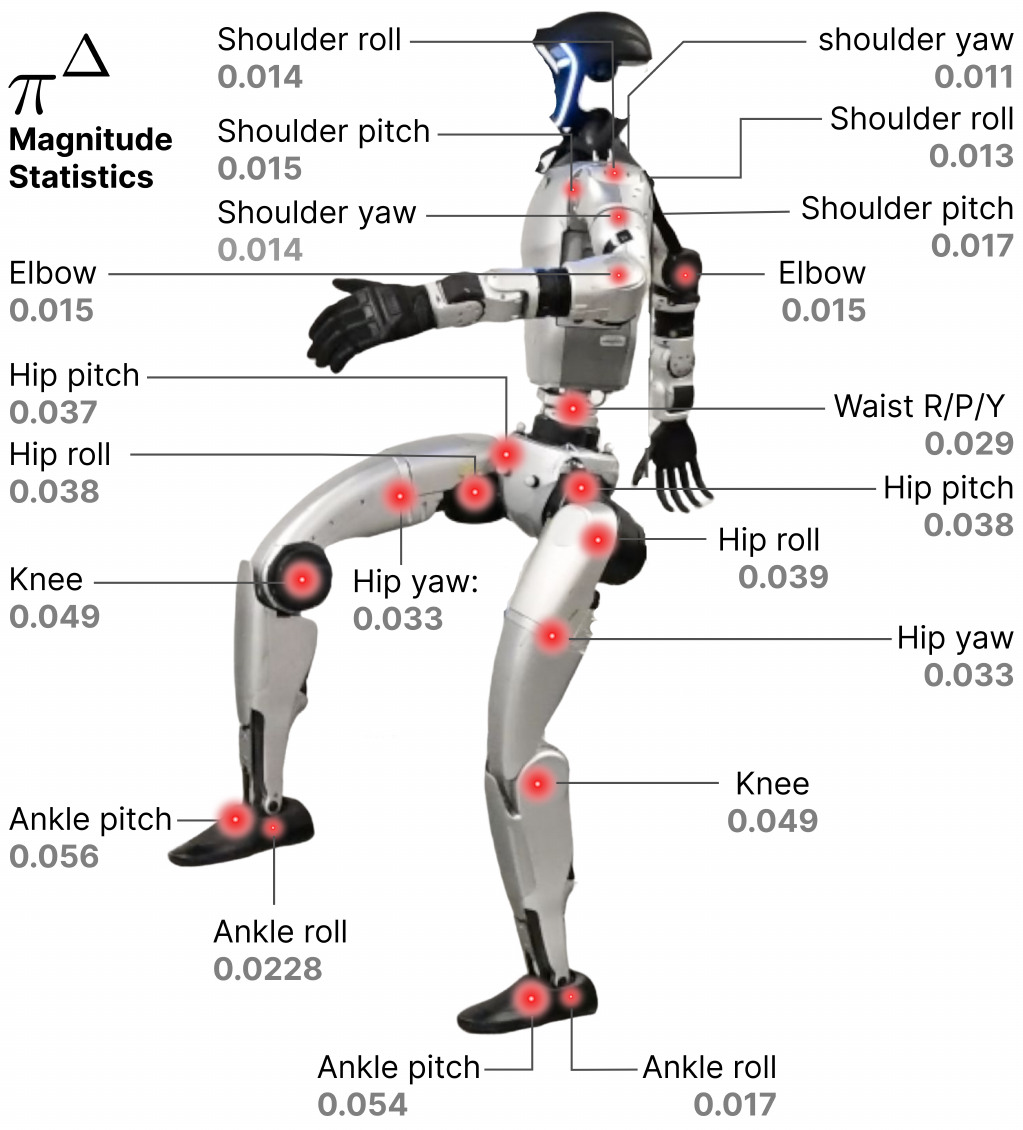}
    \vspace{-2mm}
    \caption{Visualization of IsaacGym-to-IsaacSim $\pi^\Delta$ output magnitude. We compute the average absolute value of each joint over the 4300-episode dataset. Larger red dots indicate higher values. The results suggest that lower-body motors exhibit a larger discrepancy compared to upper-body joints, with the most significant gap observed in the ankle pitch joint of the G1 humanoid.}
    \label{fig:vis_deltaA_magnitude}
    \vspace{-4mm}
\end{figure}

Random torque noise~\cite{rfi} is a widely used domain randomization technique for legged robots. To determine whether delta action facilitates fine-tuning of pre-trained policies toward real-world dynamics rather than merely enhancing robustness through random action noise, we analyze its impact. Specifically, we assess the effect of applying random action noise during policy fine-tuning in Isaac Gym by modifying the environment dynamics as $s_{t+1} = f^\text{sim}(s_t, a_t + \beta \delta_a)$, where $\delta_a \sim \mathcal{U}[0, 1]$, and deploy it in Genesis. We conduct an ablation study to examine the influence of the noise magnitude, $\beta$, varying from $0.025$ to $0.4$. As shown in~\Cref{fig:action_noise}, within the constrained range of $\beta \in [0.025, 0.2]$, policies fine-tuned with action noise outperform those without fine-tuning in terms of global tracking error (MPJPE). However, the performance of the action noise approach (MPJPE of $150$) does not match the precision achieved by \method (MPJPE of $126$). Furthermore, we visualize the average output of $\pi^\Delta$ learned from IsaacSim data in~\Cref{fig:vis_deltaA_magnitude}, which reveals non-uniform discrepancies across joints. For example, in the G1 humanoid robot under our experimental setup, lower-body motors exhibit a larger dynamics gap compared to upper-body joints. Within the lower-body, the ankle and knee joints show the most pronounced discrepancies. Additionally, asymmetries between the left and right body motors further highlight the complexity. Such structured discrepancies cannot be effectively captured by merely adding uniform action noise.
These findings, along with the results in~\Cref{fig:ASAP_openloop_curves}, demonstrate that delta action not only enhances policy robustness but also enables effective adaptation to real-world dynamics, outperforming naive randomization strategies.

\section{Related Works}
\label{sec:relatedwork}

\subsection{Learning-based Methods for Humanoid Control}

In recent years, learning-based methods have made significant progress in whole-body control for humanoid robots. Primarily leveraging reinforcement learning algorithms~\cite{schulman2017proximal} within physics simulators~\cite{makoviychuk2021isaac, mittal2023orbit, todorov2012mujoco}, humanoid robots have learned a wide range of skills, including robust locomotion~\cite{li2019using, xie2020learning, li2021reinforcement, liao2024berkeley, li2024reinforcement, radosavovic2024real, radosavovic2402humanoid, gu2024advancing, zhang2024whole}, jumping~\cite{li2023robust}, and parkour~\cite{long2024learning, zhuang2024humanoid}. More advanced capabilities, such as dancing~\cite{zhang2024wococo, ji2024exbody2, cheng2024expressive}, loco-manipulation~\cite{he2024learning, lu2024mobile, fu2024humanplus, he2024omnih2o}, and even backflipping~\cite{Unitree2024H1Backflip}, have also been demonstrated. Meanwhile, the humanoid character animation community has achieved highly expressive and agile whole-body motions in physics-based simulations~\cite{peng2022ase, tessler2024maskedmimic, luo2023perpetual}, including cartwheels~\cite{peng2018deepmimic}, backflips~\cite{peng2018sfv}, sports movements~\cite{yuan2023learning, wang2024strategy, luo2024smplolympics, wang2023physhoi, wang2024skillmimic}, and smooth object interactions~\cite{tessler2024maskedmimic, gao2024coohoi, hassan2023synthesizing}. However, transferring these highly dynamic and agile skills to real-world humanoid robots remains challenging due to the dynamics mismatch between simulation and real-world physics. 
To address this challenge, our work focuses on learning and compensating for this dynamics mismatch, enabling humanoid robots to perform expressive and agile whole-body skills in the real world. 

\subsection{Offline and Online System Identification for Robotics}

The dynamics mismatch between simulators and real-world physics can be attributed to two primary factors: inaccuracies in the robot model descriptions and the presence of complex real-world dynamics that are difficult for physics-based simulators to capture. Traditional approaches address these issues using system identification (SysID) methods~\cite{kozin1986system, aastrom1971system}, which calibrate the robot model or simulator based on real-world performance. These methods can be broadly categorized into \textit{offline SysID} and \textit{online SysID}, depending on whether system identification occurs at test time. \textit{Offline SysID} methods typically collect real-world data and adjust simulation parameters to train policies in more accurate dynamics. The calibration process may focus on modeling actuator dynamics~\cite{tan2018sim, hwangbo2019learning, yang2024agile}, refining robot dynamics models~\cite{khosla1985parameter, an1985estimation, gautier2011dynamic, han2020iterative, janot2013generic}, explicitly identifying critical simulation parameters~\cite{yu2019sim, chebotar2019closing, du2021auto,wu2024loopsr}, learning a distribution over simulation parameters~\cite{ramos2019bayessim, heiden2022probabilistic, antonova2022bayesian}, or optimizing system parameters to maximize policy performance~\cite{muratore2021data, ren2023adaptsim}. \textit{Online SysID} methods, in contrast, aim to learn a representation of the robot’s state or environment properties, enabling real-time adaptation to different conditions. These representations can be learned using optimization-based approaches~\cite{yu2018policy, yu2020learning, lee2022pi, peng2020learning}, regression-based methods~\cite{yu2017preparing, kumar2021rma, wang2024cts, gu2024advancing, ji2022concurrent, margolis2023learning, fu2023deep, qi2023hand, margolis2024rapid, kumar2022adapting, miki2022learning, lee2020learning}, next-state reconstruction techniques~\cite{nahrendra2023dreamwaq, long2024hybrid, luo2024pie, wang2024toward, shirwatkar2024pip}, direct reward maximization~\cite{li2024reinforcement}, or by leveraging tracking and prediction errors for online adaptation~\cite{o2022neural, lyu2024rl2ac, huang2023datt, gao2024neural}.
Our framework takes a different approach from traditional SysID methods by learning a residual action model that directly compensates for dynamics mismatch through corrective actions, rather than explicitly estimating system parameters.

\subsection{Residual Learning for Robotics}

Learning a residual component alongside learned or pre-defined base models has been widely used in robotics. Prior work has explored residual policy models that refine the actions of an initial controller~\cite{silver2018residual, johannink2019residual, carvalho2022residual, alakuijala2021residual, davchev2022residual, haldar2023teach, ankile2024imitation, jiang2024transic, lee2020learning}. Other approaches leverage residual components to correct inaccuracies in dynamics models~\cite{o2022neural,karnan2020reinforced,kloss2022combining,shi2021neural,he2024self} or to model residual trajectories resulting from residual actions~\cite{chi2024iterative} for achieving precise and agile motions. RGAT~\cite{karnan2020reinforced} uses a residual action model with a learned forward dynamics to refine the simulator. Our framework builds on this idea by using RL-based residual actions to align the dynamics mismatch between simulation and real-world physics, enabling agile whole-body humanoid skills.

\section{Conclusion}
\label{sec:conclusion}
We present \method, a two-stage framework that bridges the sim-to-real gap for agile humanoid control. By learning a universal delta action model to capture dynamics mismatch, \method enables policies trained in simulation to adapt seamlessly to real-world physics. Extensive experiments demonstrate significant reductions in motion tracking errors (up to 52.7\% in \simtoreal
tasks) and successful deployment of diverse agile skills—including agile jumps and kicks—on the Unitree G1 humanoid. Our work advances the frontier of \simtoreal transfer for agile whole-body control, paving the way for versatile humanoid robots in real-world applications.

\section{Limitations}
\label{sec:limitations}
While \method demonstrates promising results in bridging the sim-to-real gap for agile humanoid control, our framework has several real-world limitations that highlights critical challenges in scaling agile humanoid control to real-world:
\begin{itemize}
    \item \textbf{Hardware Constraints}: Agile whole-body motions exert significant stress on robots, leading to motor overheating and hardware failure during data collection. Two Unitree G1 robots were broken to some extent during our experiments. This bottleneck limits the scale and diversity of real-world motion sequences that can be safely collected.
    \item \textbf{Dependence on Motion Capture Systems}: Our pipeline requires MoCap setup to record real-world trajectories. This introduces practical deployment barriers in unstructured environments where MoCap setups are unavailable.
    \item \textbf{Data-Hungry Delta Action Training}: While reducing the delta action model to 4 DoF ankle joints improved sample efficiency, training the full 23 DoF model remains impractical for real-world deployment due to the large demand of required motion clips (e.g., $>$ 400 episodes in simulation for the 23 DoF delta action training).
\end{itemize}
Future directions could focus on developing damage-aware policy to mitigate hardware risks, leveraging MoCap-free alignment to eliminate the reliance on MoCap, and exploring adaptation techniques for delta action models to achieve sample-efficient few-shot alignment.

\section*{Acknowledgments}
We thank Zhenjia Xu, Yizhou Zhao, Yu Fang for help with hardware.
We thank Pulkit Goyal, Hawkeye King, Peter Varvak and Haoru Xue for help with motion capture setup. 
We thank Ziyan Xiong, Yilin Qiao and Xian Zhou for support on Genesis integration.
We thank Rui Chen, Yifan Sun, Kai Yun, Yaru Niu, Yikai Wang and Ding Zhao for help with G1 hardware. 
We thank Xuxin Cheng, Chong Zhang and Toru Lin for always being there to help with any problem.
We thank Unitree Robotics for help with G1 support.



\clearpage

\bibliographystyle{plainnat}
\bibliography{references}

\clearpage
\section*{Appendix}

\subsection{Domain Randomization in Pre-Training}
\label{sec:dr-pre-train}
To improve the robustness and generalization of the pre-trained policy in \Cref{fig:ASAP} (a), we utilized the domain randomization technics listed in \Cref{tab:deepmimic_DR}.
\begin{table}[h]
    \centering
    \setlength{\tabcolsep}{3pt} 
    \caption{Domain Randomizations}
    \begin{tabular}{cc}
        \toprule
        Term & Value \\
        \midrule
        \multicolumn{2}{c}{Dynamics Randomization} \\
        \midrule
        Friction & $\mathcal{U}(0.2,1.1)$ \\
        P Gain & $\mathcal{U}(0.925,1.05) \times$ default \\
        Control delay & $\mathcal{U}(20,40) \mathrm{ms}$ \\
        \midrule
        \multicolumn{2}{c}{External Perturbation} \\
        \midrule
        Push robot & interval $= 10 s, v_{x y} = 0.5 \mathrm{~m} / \mathrm{s}$ \\
        \bottomrule
    \end{tabular}
    \label{tab:deepmimic_DR}
\end{table}


\subsection{SysID Parameters}
We identify the following representative robot parameters in our simulated model that best align the ones in the real world: base center of mass (CoM) shift $(c_x, c_y, c_z)$, base link mass offset ratio $k_m$ and low-level PD gain ratios $(k^i_p, k^i_d)$ where $i=1,2,...,23$, as shown in~\Cref{tab:sysid_params}.
\begin{table}[htp]
    \centering
    \small 
    \setlength{\tabcolsep}{2pt} 
    \caption{SysID Parameters}
    \label{tab:sysid_params}
    \begin{tabular}{cccc}
        \toprule
        Parameter & Range & Parameter & Range \\
        \midrule
        $c_x$ & $[-0.02, 0.0, 0.02]$ & $c_y$ & $[-0.02, 0.0, 0.02]$ \\
        $c_z$ & $[-0.02, 0.0, 0.02]$ & $k_m$ & $[0.95, 1.0, 1.05]$ \\
        $k^i_p$ & $[0.95, 1.0, 1.05]$ & $k^i_d$ & $[0.95, 1.0, 1.05]$ \\
        \bottomrule
    \end{tabular}
\end{table}.

\subsection{Implementation of Delta Dynamics Learning}
\label{sec:appendix-dynamics}
Using the collected real-world trajectory, we replay the action sequence $\{a^\text{real}_0, \dots, a^\text{real}_T\}$ in simulation and record the resulting trajectory $\{s^\text{sim}_0, \dots, s^\text{sim}_T\}$. The neural dynamics model $f^\Delta_\theta$ is trained to predict the difference: 
\[
s^\text{real}_{t+1} - s^\text{sim}_{t+1} = f^\Delta_\theta(s^\text{real}_t, a^\text{real}_t), \quad \forall t.
\]

In practice, we compute the mean squared error (MSE) loss in an autoregressive setting, where the model predicts forward for $K$ steps and uses gradient descent to minimize the loss. To balance learning efficiency and stability over long horizons, we implement a schedule that gradually increases $K$ during training. Formally, the optimization objective is:
\[
\mathcal{L}= \bigg\lVert s^\text{real}_{t+K} - \underbrace{f^\text{sim} \big( \dots f^\text{sim}}_{K} (s_t, a_t) + f^\Delta_\theta(s_t, a_t), \dots, a_{t+K} \big) \bigg\rVert.
\]

After training, we freeze the residual dynamics model $f^\Delta_\theta$ and integrate it into the simulator. During each simulation step, the robot's state is updated by incorporating the delta predicted by the dynamics model. In this augmented simulation environment, we finetune the previously pretrained policy to adapt to the corrected dynamics, ensuring improved alignment with real-world behavior.

\subsection{Derivation of Training-free Methods of Using Delta Action}
\label{sec:appendix_more_deltaA_usage}
To formalize the problem, we start by assuming one-step consistency between real and simulated dynamics:
\[
f^\text{real}(s, \pi(s)) = f^\text{sim}(s, \pi(s) + \pi^\Delta(s, \pi(s))).
\]
Under this assumption, one-step matching leads to the condition:
\begin{align}
    \pi(s) &+ \pi^\Delta(s, \pi(s)) = \hat{\pi}(s), \\
    \Rightarrow \pi(s) &= \hat{\pi}(s) - \pi^\Delta(s, \pi(s)).
    \label{eq:solve_pi}
\end{align}

To solve \Cref{eq:solve_pi}, we consider:
\begin{enumerate}
    \item \textbf{Fixed-Point Iteration}: We initialize \(y_0 = \hat{\pi}(s)\) and iteratively update:
   \begin{equation}
       y_{k+1} = \hat{\pi}(s) - \pi^\Delta(s, y_k),
       \label{eq:fix_point}
   \end{equation}
   where \(y_k\) converges to a solution after \(K\) iterations.

    \item \textbf{Gradient-Based Optimization}: Define the loss function:
    \begin{equation}
       l(y) = \| y + \pi^\Delta(s, y) - \hat{\pi}(s) \|^2.
    \end{equation}
    A gradient descent method minimizes this loss to solve for \(y\).
\end{enumerate}

These methods approximate \(\pi(s)\), but suffer from OOD issues when trained on limited trajectories. RL fine-tuning, in contrast, directly optimizes \(\pi(s)\) for real-world deployment, resulting in superior performance.

\textbf{Problem of One-Step Matching.} Note that \Cref{eq:solve_pi} is derived from the one-step matching assumption (i.e., $\pi(s) + \pi^\Delta(s, \pi(s)) = \hat{\pi}(s)$). For multi-step matching, one has to differentiate through $f^\text{sim}$, which is, in general, intractable. Therefore, both fixed-point iteration and gradient-based optimization assume one-step matching. This also explains the advantages of RL-based fine-tuning: it effectively performs a gradient-free multi-step matching procedure.

\end{document}